\documentclass[sn-mathphys,Numbered]{sn-jnl}

\usepackage{graphicx}%
\usepackage{multirow}%
\usepackage{amsmath,amssymb,amsfonts}%
\usepackage{amsthm}%
\usepackage{mathrsfs}%
\usepackage[title]{appendix}%
\usepackage{xcolor}%
\usepackage{textcomp}%
\usepackage{manyfoot}%
\usepackage{booktabs}%
\usepackage{algorithm}%
\usepackage{algorithmicx}%
\usepackage{algpseudocode}%
\usepackage{listings}%
\usepackage{bm}
\usepackage{siunitx}
\usepackage{kinematikz}
\usepackage{pgfplots}
\pgfplotsset{width=10cm,compat=1.9}
\usetikzlibrary{positioning,arrows,fit}
\usetikzlibrary{backgrounds,scopes}
\definecolor{amber}{rgb}{1.0, 0.49, 0.0}
\definecolor{applegreen}{rgb}{0.55, 0.71, 0.0}
\usepackage{longtable}
\usepackage{soul}

\theoremstyle{thmstyleone}%
\theoremstyle{thmstyletwo}%

\theoremstyle{thmstylethree}%

\begin{document}

\title[A novel step-by-step procedure for the kinematic calibration of robots using a single draw-wire encoder]{A novel step-by-step procedure for the kinematic calibration of robots using a single draw-wire encoder}


\author*[1,2]{\fnm{Giovanni} \sur{Boschetti}}\email{giovanni.boschetti@unipd.it}

\author[3]{\fnm{Teresa} \sur{Sinico}}\email{teresa.sinico@phd.unipd.it}
\equalcont{These authors contributed equally to this work.}


\affil*[1]{\orgdiv{Department of Industrial Engineering (DII)}, \orgname{University of Padua}, \orgaddress{\street{Via Venezia 1}, \city{Padova}, \postcode{35131}, \country{Italy}}}

\affil[2]{\orgdiv{Department of Information Engineering (DEI)}, \orgname{University of Padua}, \orgaddress{\street{Via Giovanni Gardenigo 6b}, \city{Padova}, \postcode{35131},  \country{Italy}}}

\affil[3]{\orgdiv{Department of Management and Engineering (DTG)}, \orgname{University of Padua}, \orgaddress{\street{Stradella S. Nicola 3}, \city{Vicenza}, \postcode{36100}, \country{Italy}}}


\abstract{Robot positioning accuracy is a key factory when performing high-precision manufacturing tasks. To effectively improve the accuracy of a manipulator, often up to a value close to its repeatability, calibration plays a crucial role. In the literature, various approaches to robot calibration have been proposed, and they range considerably in the type of measurement system and identification algorithm used. Our aim was to develop a novel step-by-step kinematic calibration procedure - where the parameters are subsequently estimated one at a time - that only uses 1D distance measurement data obtained through a draw-wire encoder. To pursue this objective, we derived an analytical approach to find, for each unknown parameter, a set of calibration points where the discrepancy between the measured and predicted distances only depends on that unknown parameter. This reduces the computational burden of the identification process while potentially improving its accuracy. Simulations and experimental tests were carried out on a 6 degrees-of-freedom robot arm: the results confirmed the validity of the proposed strategy. As a result, the proposed step-by-step calibration approach represents a practical, cost-effective and computationally less demanding alternative to standard calibration approaches, making robot calibration more accessible and easier to perform.}




\maketitle

\section{Introduction}
\label{SEC:introduction}
In the last two decades, the applications of robots in high-precision manufacturing tasks such as grinding \citep{Grinding}, milling \citep{Milling}, or riveting \citep{Riveting} have expanded extensively, and so has the demand for higher precision manipulators. 

The precision of a robot can be quantified in terms of repeatability or accuracy: repeatability is defined as the precision with which the robot's end effector returns to a previously taught position, while accuracy is defined as the precision with which the robot's end effector moves to a commanded position with respect to a spatial coordinate frame. Manipulators currently used in industry are characterized by very high repeatability, but poor accuracy. The latter is rarely given by robot manufacturers and can assume a value of some millimeters.

High repeatability is the main requirement in a variety of manufacturing and handling applications where the required robot end-effector poses are manually taught by jogging the robot through a teaching pendant. Along with high repeatability, high accuracy is required in manufacturing tasks that involve offline programming, such as drilling and laser cutting, where the robot's end-effector poses are defined with respect to an absolute or relative reference frame. Indeed, robot positioning accuracy may not be adequate for performing these types of task, and conventional machine tools may still be preferred, despite their higher cost and lower flexibility. For this reason, increasing robots' accuracy is crucial for their application in flexible and reconfigurable manufacturing systems. 

The accuracy of a robot depends on the accuracy of the robot’s mathematical model in the controller: this model computes the joint angles of the robot given an end-effector pose with respect to a reference frame. The inaccurate knowledge of the geometric and non-geometric parameters of this mathematical model is the major source of the discrepancy between the actual pose of the robot's end-effector and the pose predicted by its controller. Errors associated with geometric parameters are related to the deviation between the nominal and actual dimensions of the robot’s mechanical links, misalignment of the joint axes, and incorrect joint variable offset values used to describe the manipulator’s home position \citep{GeometricErrors}. Errors associated with non-geometric parameters are caused by deformation in the mechanical components of the robot from external load and self-gravity, mechanical wear, thermal variation, sensors and servos precision, friction, and other non-linearities, including hysteresis and gear backlash \citep{Hysteresis,Thermal}. 

A practical approach to improve the robot accuracy - up to a value close to its repeatability - is to re-evaluate the parameters in the robot's mathematical model by using a calibration scheme. This procedure mainly involves four steps: modeling, measuring, identifying parameters, and implementing error compensation. 

\subsection{Modelling}
The first step of robot calibration is to derive a mathematical model that relates the robot's joint angles to its end-effector pose and takes into account the geometric and non-geometric parameters that need to be identified. A kinematic model suitable for robot calibration should meet the three principles of completeness, continuity, and minimality \citep{ModelPrinciples}. The standard Denavit Hartenberg (DH) convention \citep{DH} is widely used for kinematic modeling in robotics; however, this model does not meet the continuity condition when two consecutive joint axes are parallel. Since most industrial robots possess this feature, significant efforts have been made to solve this problem: authors either proposed to use a simplified version of the DH model to make it continuous \citep{LaserTracker1,Ballbar1} or a modified version of the standard DH model (MDH) \citep{MDH1}, adding an additional parameter to the original convention. In addition, other kinematic models that satisfy the continuity condition have been proposed, such as the S-model \cite{S-model1}, the complete and parametrically continuos (CPC) model \citep{CPC1,CPC2} and the product of exponential (POE) based model \citep{POE1,POE2}. These models meet the conditions of completeness and continuity but usually do not meet the minimality condition, i.e. some parameters are redundant; those redundant parameters must be determined and excluded from the model before identification. 
\subsection{Measuring}
Once a complete, non-singular and minimal kinematic calibration model has been derived, it is then used to compute the predicted end-effector pose based on the nominal kinematic parameter set. The predicted end-effector pose is compared with the actual end-effector pose measured by an external measurement system and an error quantity is defined. To measure the actual end-effector pose, a variety of different measurement systems can be used, which differ considerably in their cost, accuracy, ease of use, and type of data collected. In particular, the measurement systems used to calibrate a robot can be classified into complete pose measurement and partial pose measurement: a complete pose measurement of the robot’s end-effector pose consists of three position coordinates and three orientation angles, while a partial measurement of the robot’s end-effector pose consists of less than six measured values per observation (typically ranging from 3D to 1D). The most common measurement systems used for robot calibration are laser tracking systems \citep{LaserTracker1,LaserTracker2,LaserTracker3,LaserTracker4}, vision systems \citep{VisionSystem1,VisionSystem2}, ballbars \citep{Ballbar1,Ballbar2}, theodolites \citep{Theodolites1,Theodolites2,Theodolites3}, and coordinate measuring machines (CMM) \citep{CMM1,CMM2}.
\subsection{Identifying error parameters}
When a sufficient number of end-effector poses have been measured, the unknown parameters can be estimated. This identification can be achieved by determining the analytical relationship between the end-effector coordinates and the parameters in the form of a Jacobian and then inverting the equation to calculate the deviation of the parameters from their nominal values. Alternatively, the identification can also be viewed as a constrained non-linear optimization problem. In this case, a cost function that relates the parameter to a quantity that is an overall measure of optimality is defined, and the parameter set is systematically changed to reduce the cost function to zero. This problem can be solved using different optimization algorithms, such as Levenberg-Marquardt \citep{LM1}, extended Kalman filter (EKF) \citep{EKF1,EKF2}, or particle swarm optimization (PSO) \citep{PSO1,PSO2}.
\subsection{Implementing error compensation}
The final step of robot calibration involves implementing error compensation: this can be done by modifying the nominal parameters embedded in the robot controller or by using error compensation techniques. Since most robot controllers use DH parameters, the former approach is typically possible only if the standard DH convention is used in the modeling phase and if the identification is limited to kinematic parameters. However, access to modify the kinematic parameters is not always possible and may be limited (i.e., not all parameters can be modified). On the other hand, error compensation techniques need to be implemented if a different kinematic model is used in the modeling step, if access to modify some or all the kinematic parameters is not possible, and also if the identification is extended to non-kinematic parameters. In this case, the pose deviation is first calculated using the mathematical model of the manipulator with the estimated parameters and then a compensated pose is obtained \citep{ErrorCompensation1,ErrorCompensation2}. 
\subsection{Scope and contribution}
In this paper, we present a novel step-by-step procedure for the kinematic calibration of robots using a single draw-wire encoder. We were interested in developing a kinematic calibration procedure with this measurement system because it offers a good balance of accuracy, resolution, measurement range, cost, and usability. In particular, it is much less expensive and requires an easier set-up than laser trackers, which are still the most used measurement instrument for the calibration of robots. Several studies have used multiple draw-wire encoders for robot calibration: they can be arranged as complete pose measurement devices \citep{Multiple-Encoders1,Multiple-Encoders2,Multiple-Encoders3} or partial pose measurement devices \citep{Multiple-Encoders4,Multiple-Encoders5,Multiple-Single-Encoder}. However, few studies have considered the possibility of using a single draw-wire encoder \citep{Multiple-Single-Encoder,Single-Encoder-1,Single-Encoder-2,Single-Encoder-3,Single-Encoder-4,Single-Encoder-5}, which only provides a 1D radial measurement of a point with respect to a fixed reference frame. In this case, the identification step is typically viewed as an optimization problem, where the cost function is the sum of the difference between the measured and predicted distance of the robot's end-effector from a fixed point over a number of calibration poses.  

When calibrating a robot using a single draw-wire encoder, two practical choices have to be made: the encoder location and the set of calibration points where measurements are taken. These two aspects greatly influence the resulting accuracy after calibration, but previous studies did not explore them in depth. In \cite{Multiple-Single-Encoder} the sensor is located outside the manipulator workspace, while in \citep{Single-Encoder-3} it is located in an arbitrary place: in both cases, the location of the measurement system is an unknown parameter that is found through the optimization algorithm along with the other kinematic parameters. Unfortunately, in \citep{Single-Encoder-1,Single-Encoder-2,Single-Encoder-4,Single-Encoder-5} the authors did not provide complete details of sensor location. Moreover, in \cite{Multiple-Single-Encoder} the calibration points have been chosen from a uniformly sampled grid that covers most of the working envelope in front of the robot, while in \citep{Single-Encoder-1,Single-Encoder-2,Single-Encoder-3,Single-Encoder-4,Single-Encoder-5} the details about the calibration points are not reported. 

Our step-by-step calibration procedure differs significantly from previously proposed approaches using a single draw-wire encoder, and the heart of our approach lies in the careful selection of encoder location and calibration points. The draw-wire encoder is placed in a position in the robot's workspace specified by a set of joint coordinates: this allows the estimation of kinematic parameters without explicit knowledge of the coordinates of the encoder location. This approach is more robust than the approaches proposed in \citep{Multiple-Single-Encoder,Single-Encoder-3}, as the identification of the fixed location of the encoder could introduce errors that could propagate to the final results. In addition, we developed a novel analytical approach to find a set of calibration points for each parameter that needs to be identified: in this set of points, the discrepancy between predicted and measured distances depends only on that parameter (and possibly on those that have already been identified). Consequently, our calibration procedure can be viewed as a step-by-step procedure through which the unknown parameters are subsequently identified: the manipulator is first moved to the first set of calibration points, where the measured distance error depends only on one of the kinematic parameters. This unknown parameter is identified, and the manipulator is moved to a second set of calibration points, where the error depends on a second kinematic parameter (and possibly also on the just identified parameter). The second kinematic parameter is then estimated, and, following the same procedure, all kinematic parameters are identified. The proposed approach offers two major advantages: only one of the unknown parameters is estimated at a time, and fewer measurements are processed at the same time. The consequence of the former is that the parameters can be estimated more accurately than with previously proposed approaches, and their identification does not require complex optimization algorithms; the consequence of the latter is that the measurement data can be computationally processed faster. As a result, the proposed calibration procedure can be implemented directly into the robot's controllers, which typically have limited computational power, without the need for extra hardware to process the measurement data. 
\subsection{Paper structure}
The remainder of this paper is organized as follows: in Section \ref{SEC:calibrationapproach} our novel step-by-step calibration approach is presented. The materials that we used in our experiments to test and validate the calibration procedure are described in Section \ref{SEC:materials}. The experimental results and related discussion are presented in Section \ref{SEC:results} and Section 
\ref{SEC:discussion} respectively. Finally, in Section \ref{SEC:conclusions} our conclusions are drawn. 

\section{Proposed kinematic calibration approach}
\label{SEC:calibrationapproach}
The proposed calibration procedure requires the measurement of the distance between the end-effector and a fixed location, for several poses of the robot. For this purpose, we used a draw-wire encoder fixed to the table on which the robot stands; the other end of the wire is attached to the robot's flange by a specifically designed fixture. If the fixed location of the draw-wire encoder is known, the distance between the end-effector and the draw-wire encoder exit point at each calibration point can be easily computed as
\begin{equation}
    d_i=\sqrt{\Delta x_{i-0}^2+\Delta y_{i-0}^2 +\Delta z_{i-0}^2} =\sqrt{(x_i-x_0)^2+(y_i-y_0)^2+(z_i-z_0)^2},
    \label{EQ:distance1}
\end{equation}
where $(x_i,y_i,z_i)$ are the coordinates of the $i$th calibration point and $(x_0,y_0,z_0)$ are the coordinates of the fixed location of the draw-wire encoder. For a $n$ degrees-of-freedom (DOF) manipulator, the coordinates of the $i$th calibration point with respect to the robot's base frame can be computed through the direct kinematics, which is a nonlinear function of the manipulator's joint variables and geometric parameters
\begin{equation}
    \begin{cases}
    x_i=f_x(\bm{\theta_i},\bm{p}) \\
    y_i=f_y(\bm{\theta_i},\bm{p}) \\
    z_i=f_z(\bm{\theta_i},\bm{p}) 
    \end{cases},
    \label{EQ:coordinates1}
\end{equation}
where $\bm{\theta_i}=[\theta_i,\dots,\theta_n]$ are the joint coordinates of the robot at the $i$th calibration point and $\bm{p}=[p_1,\dots,p_j]$ is the vector containing the geometric parameters of the manipulator, which depend on the model chosen to describe the robot's kinematics. At each calibration point, the computed distance will be different from the measured distance, due to the discrepancy between the nominal and actual values of the model parameters. We focused on identifying only kinematic parameters, neglecting non-kinematic parameters, as inaccurate knowledge of kinematic parameters yields about 90\% of the total positioning inaccuracy \citep{90-1,Theodolites2}. In particular, both the joint coordinates $\bm{\theta}$ and the geometric parameters $\bm{p}$ are affected by errors, which can be seen as offsets from their nominal values. At this stage, we need to define the set of kinematic error parameters that we wish to estimate (for example, we may only be interested in identifying joint offsets). The model parameters can be grouped in vector $\bm{e}=[e_1,\dots,e_m]$, while their offsets from their nominal value can be grouped in the error vector $\bm{\delta e}=[\delta e_1,\dots,\delta e_m]$. Given the nominal geometric parameters of the manipulator $\bm{p_{n}}$ and the joint coordinates at the $i$th calibration point $\bm{\theta_i}$, the spatial coordinates in Equation \eqref{EQ:coordinates1} can be rewritten as a function of the error vector $\bm{\delta e}$ and the distance in Equation \eqref{EQ:distance1} can be rewritten as
\begin{equation}
    d_i(\bm{\delta e})=\sqrt{(x_i(\bm{\delta e})-x_0)^2+(y_i(\bm{\delta e})-y_0)^2+(z_i(\bm{\delta e})-z_0)^2}
    \label{EQ:distance2}.
\end{equation}
The difference between the computed $d_i(\bm{e})$ and measured distance $\tilde{d}_i$, known as the residual, is calculated at each calibration point. The aggregate sum of squares of the residuals over $N$ calibration points is then calculated as
\begin{equation}
    f_{cost}(\bm{\delta e})=\sum_{i=1}^N (d_i(\bm{\delta e}) - \tilde{d}_i)^2.
    \label{EQ:costfunction}
\end{equation}
Calibration may be posed as the problem of systematically varying the error parameters $\bm{e}$ in order to minimize the error in wire length over the set of measurement points: the quantity in Equation \ref{EQ:costfunction} is used as a cost function to be minimized by error parameter estimation. The optimal error vector $\bm{\delta e^*}$ is calculated as
\begin{equation}
    \bm{\delta e^*}=\arg \min_{\bm{\delta e} \in \bm{\Omega}} f_{cost}(\bm{\delta e}),
\end{equation}
where $\bm{\Omega}$ is the set of allowable deviations of the error parameters. This is a non-linear least-squares optimization problem that can be solved using various optimization algorithms. 

\subsection{Encoder location} 
\begin{figure}
    \centering
    \includegraphics[width=\textwidth]{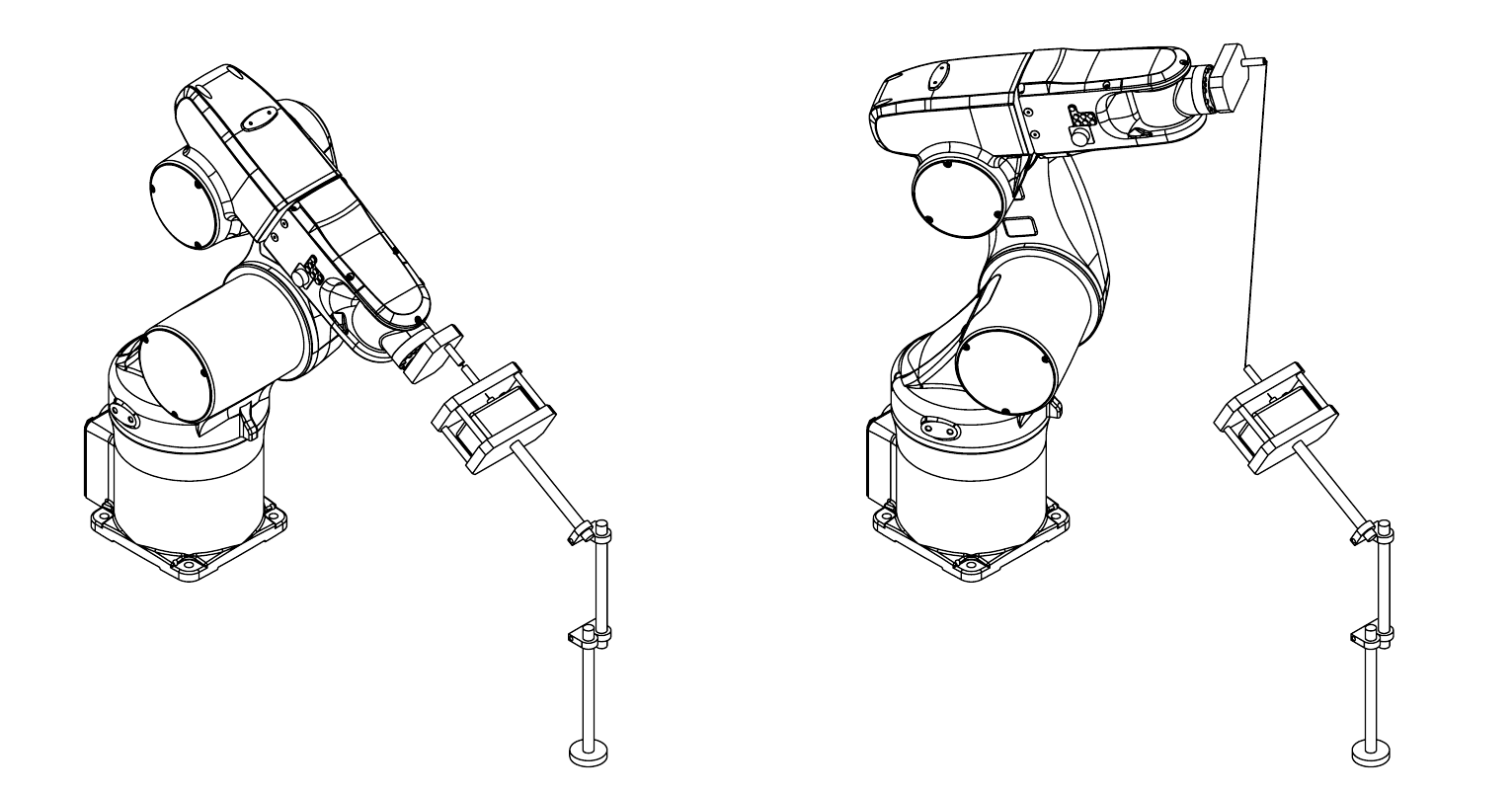}
    \caption{System at the configuration specified by the joint coordinates $\bm{\theta_0}$ (left) and at the $i$th calibration configuration, specified by the joint coordinates $\bm{\theta_i}$ (right).}
    \label{FIG:encoderlocation}
\end{figure}
Equation \ref{EQ:distance2} requires the precise knowledge of the encoder location with respect to the robot's base frame, which is not easy to obtain. In previous studies \citep{Multiple-Single-Encoder,Single-Encoder-3}, encoder location coordinates were treated as additional unknown parameters that were identified by the optimization algorithm after an initial estimate. In the work proposed here, we first moved the robot to the configuration specified by the joint coordinates $\bm{\theta_0}$ and then mounted the draw wire encoder so that the two tips touch when the robot is in this configuration (i.e., the measured distance is zero), as depicted in Figure \ref{FIG:encoderlocation}. In this case, $(x_0,y_0,z_0)$ are not independent additional parameters, but they depend on the manipulator's parameters and joint variables through the direct kinematics as well
\begin{equation}
    \begin{cases}
    x_0=f_x(\bm{\theta_0},\bm{p}) \\
    y_0=f_y(\bm{\theta_0},\bm{p}) \\
    z_0=f_z(\bm{\theta_0},\bm{p}) 
    \end{cases}.
    \label{EQ:coordinates0}
\end{equation}
However,the actual encoder coordinates are not exactly the ones calculated through Equation \eqref{EQ:coordinates0}, as the joint coordinates $\bm{\theta}$ and geometric parameters $\bm{p}$ are affected by errors. Given the nominal geometric parameters of the manipulator $\bm{p_{n}}$ and the joint coordinates $\bm{\theta_0}$, the spatial coordinates in \eqref{EQ:coordinates0} can also be written as functions of the error vector $\bm{\delta e}$. Therefore, the computed distance expressed in Equation \ref{EQ:distance2} can be rewritten as
\begin{equation}
    d_i(\bm{\delta e})=\sqrt{(x_i(\bm{\delta e})-x_0(\bm{\delta e}))^2+(y_i(\bm{\delta e})-y_0(\bm{\delta e}))^2+(z_i(\bm{\delta e})-z_0(\bm{\delta e}))^2}.
    \label{EQ:distance3}
\end{equation}
Both the $i$th calibration point and encoder location coordinates are written as a function of the same set of error parameters $\bm{e}$, without the need for three additional parameters. 

\subsection{Calibration points}
Once we defined the cost function and the location of the encoder, we were interested in finding an optimal set of calibration points, which is a well-known issue of robot calibration. In fact, the resulting accuracy after calibration is strongly dependent on the selection of the measurement poses \citep{Observability1,Observability2}. Depending on the selected pose in the robot's workspace, a variation in one of the error parameters could produce a small or large error on the end-effector pose. Ideally, a variation in any of the error parameters should cause the maximum possible error on the gripper pose so that the effect of noise (due to unmodeled error sources and measurement errors) can be minimized and, consequently, all of the error parameters can be accurately identified. To measure the goodness of a set of calibration points, different observability indices \citep{Observability3}, which are based on the singular value decomposition (SVD) of the Jacobian identification, have been proposed. Unfortunately, this approach is not applicable in our case, as we do not measure the end-effector pose directly, and it is not possible to derive a Jacobian matrix relative to the distance error. Both the coordinates of the $i$th calibration point and of the fixed location of the encoder depend on the direct kinematics of the manipulator; these coordinates appear in Equation \eqref{EQ:distance3} subtracted from each other, then squared and added. For this reason, rather than evaluating the contribution of the different error components on the end-effector pose, we evaluated their contribution on the distance from a fixed location specified in the joint space. 

First, to understand the influence of the different error parameters at the $i$th calibration point, we can calculate the partial derivatives of \eqref{EQ:coordinates1} with respect to the vector $\bm{e}$
\begin{equation}
    \bm{\Phi_i}=\begin{bmatrix}
        \partial f_x/\partial \bm{e} \\
        \partial f_y/\partial \bm{e} \\
        \partial f_z/\partial \bm{e} \\
    \end{bmatrix}_{\substack{\bm{\theta}=\bm{\theta_i}\\ \bm{p}=\bm{p_n}}}=\begin{bmatrix}
            \partial f_x/\partial e_1 & \dots & \partial f_x/\partial e_m \\
            \vdots & \ddots & \vdots \\
            \partial f_z/\partial e_1 & \dots & \partial f_z/\partial e_m 
        \end{bmatrix}_{\substack{\bm{\theta}=\bm{\theta_i}\\ \bm{p}=\bm{p_n}}}  .
\end{equation}
This is a $3\times m$ matrix (corresponding to half of the analytic Jacobian of the robot), where $m$ is the number of unknown parameters that we are trying to estimate, and it is evaluated at nominal conditions, i.e., $\bm{\theta}=\bm{\theta_i}$ and $\bm{p}=\bm{p_{n}}$. Similarly, we can calculate the partial derivatives of \eqref{EQ:coordinates0} with respect to the vector $\bm{e}$
\begin{equation}
    \bm{\Phi_0}=\begin{bmatrix}
        \partial f_x/\partial \bm{e} \\
        \partial f_y/\partial \bm{e} \\
        \partial f_z/\partial \bm{e} \\
    \end{bmatrix}_{\substack{\bm{\theta}=\bm{\theta_0}\\ \bm{p}=\bm{p_n}}}=\begin{bmatrix}
            \partial f_x/\partial e_1 & \dots & \partial f_x/\partial e_m \\
            \vdots & \ddots & \vdots \\
            \partial f_z/\partial e_1 & \dots & \partial f_z/\partial e_m 
        \end{bmatrix}_{\substack{\bm{\theta}=\bm{\theta_0}\\ \bm{p}=\bm{p_n}}}.
\end{equation}
In Equation \eqref{EQ:distance3} the coordinates of the $i$th calibration point and the coordinates of the encoder location appear subtracted from each other. The contribution of the different error parameters on the quantities $\Delta x_{i-0}$, $\Delta y_{i-0}$ and $\Delta z_{i-0}$ that appear in \eqref{EQ:distance1} is given by
\begin{equation}
    \bm{\Phi_{i-0}}=\bm{\Phi_i}-\bm{\Phi_0}.
\end{equation}
The draw-wire encoder only supplies a 1D radial measurement of the wire: rather than the influence of the different parameters on $\Delta x_{i-0}$, $\Delta y_{i-0}$ and $\Delta z_{i-0}$, we are interested in the influence of the different parameters on the direction of the wire. For this purpose, let us define
\begin{equation}
    \begin{cases}
        \Delta x_{i-0,th}=f_x(\bm{\theta_i},\bm{p_n})-f_x(\bm{\theta_0},\bm{p_n}) \\
        \Delta y_{i-0,th}=f_y(\bm{\theta_i},\bm{p_n})-f_y(\bm{\theta_0},\bm{p_n}) \\
        \Delta z_{i-0,th}=f_y(\bm{\theta_i},\bm{p_n})-f_y(\bm{\theta_0},\bm{p_n}) 
    \end{cases},
\end{equation}
i.e. the theoretical values of $\Delta x_{i-0}$, $\Delta y_{i-0}$ and $\Delta z_{i-0}$ (evaluated with nominal parameters). With these quantities, we can define the vector 
\begin{equation}
    \bm{\nu_i}=\begin{bmatrix}
        \Delta x_{i-0,th} & \Delta y_{i-0,th} & \Delta z_{i-0,th}
    \end{bmatrix},
\end{equation}
and the corresponding unit vector $\bm{\hat{\nu}}_i$, which represents the theoretical direction of the wire at the $i$th calibration point
\begin{equation}
    \bm{\hat{\nu}}_i=\frac{\bm{\nu_i}}{\| \bm{\nu_i} \|}.
\end{equation}
If the error vector caused by one error component is nearly perpendicular to the direction of the wire, this error component will have little influence on the residual that appears in Equation \ref{EQ:costfunction}. On the other hand, if the error vector caused by that error component is parallel to the direction of the wire, this error component will have a great influence on the residual. For this reason, to evaluate the contribution of the $j$th error component at the $i$th calibration point, we can compute the projection of the $j$th column of $\bm{\Phi_{i-0}}$ in the direction of the theoretical distance of the wire 
\begin{equation}
    \Psi_{i_j} = \langle \bm{\Phi_{{i-0}_j}},\bm{\hat{\nu}}_i \rangle .
\end{equation}
Consequently, the influence of the different error parameters $\bm{e}$ on the wire length can be expressed by the following matrix
\begin{equation}
    \bm{\Psi_i} = \begin{bmatrix}
        \Psi_{i_1} & \dots & \Psi_{i_m}
    \end{bmatrix} =
    \begin{bmatrix}
        \langle \bm{\Phi_{{i-0}_1}},\bm{\hat{\nu}}_i  \rangle & \dots & \langle \bm{\Phi_{{i-0}_m}},\bm{\hat{\nu}}_i  \rangle
    \end{bmatrix}.
    \label{EQ:Psi}
\end{equation} 
This matrix allows the evaluation of the contribution of each error component on the residual at a definite pose in the robot's workspace: the higher each component of $\bm{\Psi_i}$, the higher the contribution of the related error component on the wire length. When analyzing the matrix in Equation \eqref{EQ:Psi} it is important to properly compare terms related to angular errors and terms related to linear errors. If $\Psi_j$ is related to an angular error, it is expressed in $\unit{\milli\meter / \radian}$, while if $\Psi_k$ is related to a linear error, it is dimensionless. Considering that angular errors are typically in the range of $\SI{\pm 1}{\degree}$ while linear errors are in the range of $\SI{\pm 1}{\milli\meter}$, we can convert $\Psi_j$ in $\unit{\milli\meter / \degree}$: in this way, both $\Psi_j$ and $\Psi_k$ multiply a term that is within the range $\pm 1$ of their respective measurement units. 

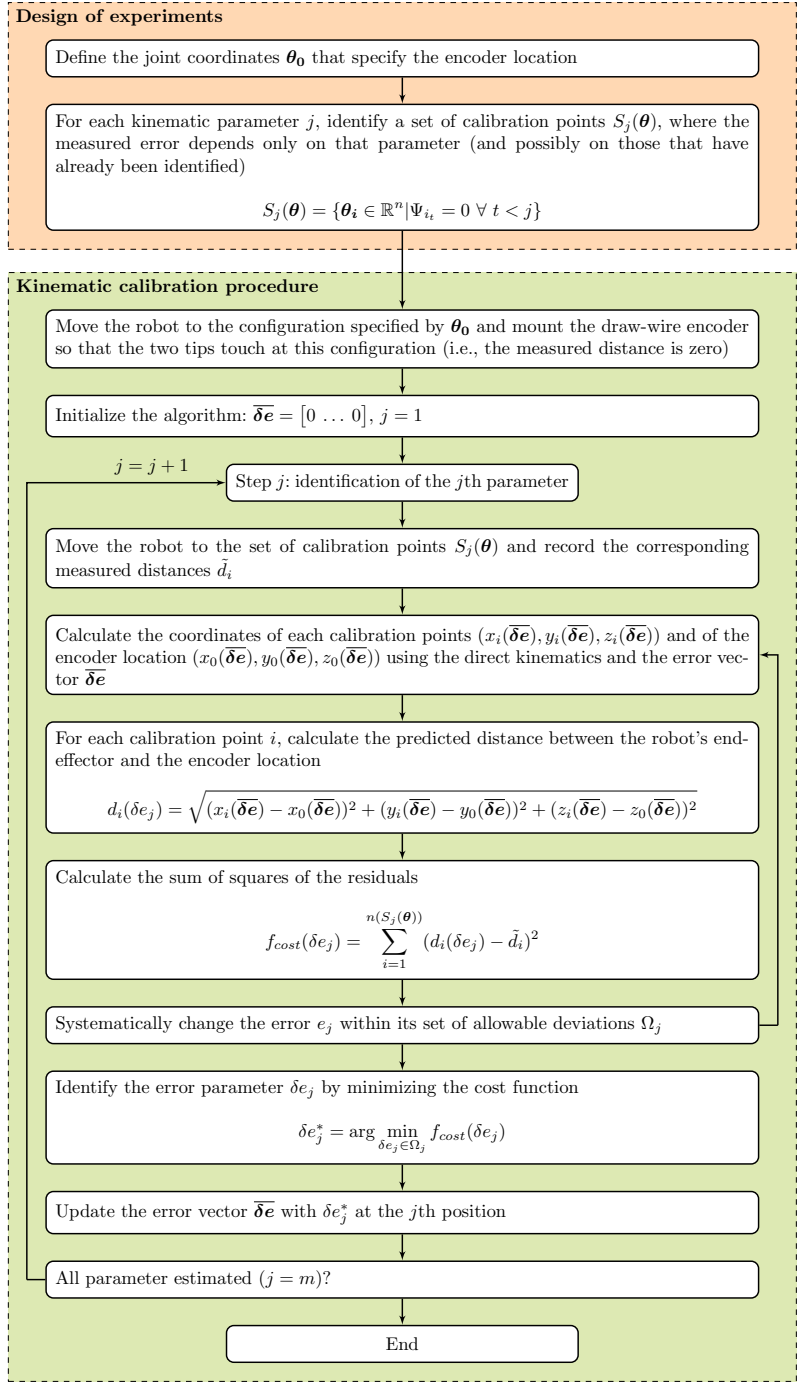
\begin{figure}
    \centering
    \resizebox{0.8\textwidth}{!}{%
    \begin{tikzpicture}[
    box/.style={rectangle,draw,fill=white,node distance=1cm,text width=37.5em,align=justify,rounded corners,minimum height=2em,thick,inner xsep=5pt,inner ysep=5pt},
    arrow/.style={draw,-latex',thick},
  ]

  \node [box] (theta0) {Define the joint coordinates $\bm{\theta_0}$ that specify the encoder location};

  \node [box, below=0.5 of theta0] (set-of-points) {For each kinematic parameter $j$, identify a set of calibration points $S_j(\bm{\theta})$, where the measured error depends only on that parameter (and possibly on those that have already been identified)
  \begin{equation*}
      S_j(\bm{\theta})=\{ \bm{\theta_i} \in \mathbb{R}^n | \Psi_{i_t}=0 \ \forall \ t < j \}
  \end{equation*}
  };
  \node [box,below=1.5 of set-of-points] (move-encoder) {Move the robot to the configuration specified by $\bm{\theta_0}$ and mount the draw-wire encoder so that the two tips touch at this configuration (i.e., the measured distance is zero)};
  \node [box,below=0.5 of move-encoder] (initialize) {Initialize the algorithm: $\bm{\overline{\delta e}}=\begin{bmatrix} 0 & \dots & 0 \end{bmatrix}$, $j=1$};
  \node [box,below=0.5 of initialize,text width=18em] (step-j) {Step $j$: identification of the $j$th parameter};
  \node [box,below=0.5 of step-j] (move-robot) {Move the robot to the set of calibration points $S_j(\bm{\theta})$ and record the corresponding measured distances $\tilde{d}_i$};
  \node [box,below=0.5 of move-robot] (calc-coord) {Calculate the coordinates of each calibration points $(x_i(\bm{\overline{\delta e}}),y_i(\bm{\overline{\delta e}}),z_i(\bm{\overline{\delta e}}))$ and of the encoder location $(x_0(\bm{\overline{\delta e}}),y_0(\bm{\overline{\delta e}}),z_0(\bm{\overline{\delta e}}))$ using the direct kinematics and the error vector $\bm{\overline{\delta e}}$};

    \node [box,below=0.5 of calc-coord] (calc-dist) {For each calibration point $i$, calculate the predicted distance between the robot's end-effector and the encoder location 
    \begin{equation*}
        d_i(\delta e_j)=\sqrt{(x_i(\bm{\overline{\delta e}})-x_0(\bm{\overline{\delta e}}))^2+(y_i(\bm{\overline{\delta e}})-y_0(\bm{\overline{\delta e}}))^2+(z_i(\bm{\overline{\delta e}})-z_0(\bm{\overline{\delta e}}))^2}
    \end{equation*}
    };

\node [box,below=0.5 of calc-dist] (calc-res) {Calculate the sum of squares of the residuals
    \begin{equation*}
        f_{cost}(\delta e_j)=\sum_{i=1}^{n(S_j(\bm{\theta}))} (d_i(\delta e_j)-\tilde{d}_i)^2
    \end{equation*}
    };

\node [box,below=0.5 of calc-res] (ch-parameter) {Systematically change the error $e_j$ within its set of allowable deviations $\Omega_j$
    };

\node [box,below=0.5 of ch-parameter] (det-parameter) {Identify the error parameter $\delta e_j$ by minimizing the cost function
\begin{equation*}
    \delta e_j^*=\arg \min_{\delta e_j \in \Omega_j} f_{cost} (\delta e_j)
\end{equation*}
};

\node [box,below=0.5 of det-parameter] (up-error) {Update the error vector $\bm{\overline{\delta e}}$ with $\delta e_j^*$ at the $j$th position
};

\node [box,below=0.5 of up-error] (fine?) {All parameter estimated ($j=m$)?};

\node [box,below=0.5 of fine?,text width=18em, align=center] (end) {End};

\path
  (ch-parameter.east) ++(1em,0em) coordinate (prova);

\path
  (fine?.west) ++(-1em,0em) coordinate (prova2);

\path
  (step-j.west) ++(-4em,0.1em) coordinate (prova3);

\draw (prova3) node[anchor=south] {$j=j+1$};

    \path
  (move-encoder.north west) ++(-1em,1em) coordinate (move fit)
  (end.south east) ++(10.75em,-0em) coordinate (up error fit);

  \path
  (theta0.north west) ++(-1em,1em) coordinate (theta0 fit)
  (set-of-points.south east) ++(1em,-0em) coordinate (set of points fit);

  \draw (theta0 fit) ++(5em,1em) node[anchor=north] {\textbf{Design of experiments}};

  \draw (move fit) ++(7.6em,1em) node[anchor=north] {\textbf{Kinematic calibration procedure}};

  \path [arrow] (theta0) -- (set-of-points);
  \path [arrow] (set-of-points) -- (move-encoder);
  \path [arrow] (move-encoder) -- (initialize);
  \path [arrow] (initialize) -- (step-j);
  \path [arrow] (step-j) -- (move-robot);
  \path [arrow] (move-robot) -- (calc-coord);
  \path [arrow] (calc-coord) -- (calc-dist);
  \path [arrow] (calc-dist) -- (calc-res);
  \path [arrow] (calc-res) -- (ch-parameter);
  \path [arrow] (ch-parameter) -- (det-parameter);
  \path [arrow] (det-parameter) -- (up-error);
  \path [arrow] (up-error) -- (fine?);
  \path [arrow] (fine?) -- (end);

  \path [draw,thick] (ch-parameter.east) -- (prova);
  \draw [arrow] (prova) |- (calc-coord.east);

  \path [draw,thick] (fine?.west) -- (prova2);
  \draw [arrow] (prova2) |- (step-j.west);

  \node [rectangle,draw,dashed,inner sep=1em,fit=(move fit) (up error fit)] (enclosure1) {};

  \node [rectangle,draw,dashed,inner sep=1em,fit=(theta0 fit) (set of points fit)] (enclosure2) {};

  {[on background layer] 
        \fill[even odd rule,amber!30] (enclosure2.south west) rectangle (enclosure2.north east);
        \fill[even odd rule,applegreen!30] (enclosure1.south west) rectangle (enclosure1.north east);
    }


\end{tikzpicture}
}
    \caption{Flowchart of the proposed step-by-step calibration procedure}
    \label{FIG:flowchart}
\end{figure}

The matrix $\bm{\Psi_i}$ not only allows the evaluation of the contributions of different error parameters on the residual at a specific point in the workspace but also gives us a key insight: since the position of the draw-wire encoder is specified by the joint coordinates $\bm{\theta_0}$, by appropriately choosing the joint coordinates $\bm{\theta_i}$, some of the terms in $\bm{\Psi_i}$ may become null. In particular, we can find a set of joint coordinates $\bm{\theta_0}$ and $\bm{\theta_i}$ so that
\begin{equation}
    (\bm{\theta_i},\bm{\theta_0})=\{ \bm{\theta_i},\bm{\theta_0} \in  \mathbb{R}^n | \Psi_{i_j} \neq 0, \Psi_{i_k}=0 \ \forall \ k \neq j \}.
    \label{EQ:property}
\end{equation}
This means that all the elements of $\bm{\Psi_i}$ are null except for $\Psi_{i_j}$ and this is possible if some of the joint coordinates between $\bm{\theta_0}$ and $\bm{\theta_i}$ are kept constant while some others are changed (typically the last two in the case of a 6 DOF anthropomorphic robot arm). In this case, the difference between the predicted and measured distance depends only on the $j$th error parameter: the cost function becomes a function of $\delta e_j$ only, and the remaining parameters are kept constant at their nominal values. By choosing a proper set of calibration points $S_j(\bm{\theta})$ that satisfy Equation \eqref{EQ:property}, the $j$th error parameter can be accurately estimated by minimizing the cost function as:
\begin{equation}
    \delta e_j^* = \arg \min_{\delta e_j \in \Omega_j} f_{cost}(\delta e_j).
    \label{EQ:ej}
\end{equation}
After identifying the $j$th error parameter, we can find a second set of calibration points $S_w(\bm{\theta})$ where all the elements of $\bm{\Psi_i}$ are null except for $\Psi_{i_j}$ and $\Psi_{i_w}$: this allows the estimation of the $w$th parameter. In fact, the cost function becomes a function of $\delta e_w$ only, while the remaining parameters are kept constant ($\delta e_j$ is kept at the value found at the previous step with Equation \eqref{EQ:ej} and the remaining parameters are kept at their nominal value). Following this procedure and finding different sets of points with these properties, we can estimate the remaining error parameters, one at a time. In addition, once we determine the order in which the parameters will be identified,  $\bm{e}$ can be rearranged in vector $\bm{\overline{e}}$ and $\bm{\Psi_i}$ in $\bm{\overline{\Psi}_i}$, so that the order of the columns reflects the order in which the parameters are estimated. For each unknown parameter $j$, the corresponding set of calibration points can be expressed as:
\begin{equation}
    S_j(\bm{\theta})= \{ \bm{\theta_i} \in \mathbb{R}^n | \Psi_{i_t} = 0 \ \forall \ t > j \}.
    \label{EQ:pointsets}
\end{equation}
As a result, we propose the step-by-step kinematic calibration procedure described in Figure \ref{FIG:flowchart}, through which the error parameters are subsequently identified, one at a time. This constitutes a major advantage over standard calibration approaches, where the measured error depends on all the parameters and the unknown parameters are estimated all together, often with complex optimization algorithms. In fact, if the measured error depends on all the error parameters, it is more difficult to accurately estimate each of them, and the optimization algorithm may converge to a set of error parameters that minimize the cost function, but do not represent the actual error parameter values. 

The validity of the proposed step-by-step calibration procedure can also be demonstrated with a different approach, described in \ref{APP:1}.

\section{Materials and methods}
\label{SEC:materials}
\subsection{Industrial manipulator: the Adept Viper S650}
To test the proposed calibration approach, we used the Adept Viper S650 robot and the eMB-60R controller, both manufactured by Omron Adept Inc. The Adept Viper S650 is an articulated robot with six degrees of freedom and has a repeatability of $\SI{\pm 0.02}{\milli\meter}$. The robot controller uses DH parameters, and the DH table according to the convention described in \citep{Craig} is presented in Table \ref{TAB:DHtable}. 
\begin{table}[]
    \caption{DH parameters of the Adept Viper S650 manipulator.}
    \label{TAB:DHtable}
    \centering
    \begin{tabular}{ccccc}
    \toprule
    $i$ & $\alpha_{i-1}$ & $a_{i-1}$ & $\theta_i$ & $d_i$ \\
    & $(\unit{\degree})$ & $(\unit{\milli\meter})$ & $(\unit{\degree})$ & $(\unit{\milli\meter})$ \\
    \midrule
    $1$ & $0$ & $0$ & $\theta_1$ & $0$ \\
    $2$ & $-90$ & $75$ & $\theta_2$ & $0$ \\
    $3$ & $0$ & $270$ & $\theta_3$ & $0$ \\
    $4$ & $90$ & $-90$ & $\theta_4$ & $295$ \\
    $5$ & $-90$ & $0$ & $\theta_5$ & $0$ \\
    $6$ & $90$ & $0$ & $\theta_6$ & $80$ \\
    \bottomrule
    \end{tabular}
\end{table}
\subsection{Measurement system: draw-wire encoder}
The measurement system used to obtain 1D distance measurements consists of a draw-wire encoder manufactured by Unimeasure Inc. (model EP-50-N20-N30-10C) and a specifically designed tool to be attached directly to the robot's flange. The draw-wire encoder is composed of a reel on which the wire is wound, an incremental encoder, and electronic components for signal conditioning. The incremental encoder measures the rotation of the reel, and hence the amount of extracted wire, while the spring maintains the proper tension cable. The main specifications of this measuring system are a resolution of $\SI{0.025488}{\milli\meter}$ and a maximum measured length of $\SI{1250}{\milli\meter}$, which is suitable for our experiments, as the manipulator's workspace is approximately a hemisphere with a radius of about $\SI{600}{\milli\meter}$. The draw-wire encoder can be easily placed in different positions and orientations, thanks to its holding fixture. In addition, the draw-wire encoder is connected to the robot controller through the belt encoder port (which is typically used for conveyor tracking tasks). Measurement data are acquired directly from the robot controller, which then proceeds to process the data and identify error parameters. 
\subsection{Kinematic calibration model}
Since we wanted our kinematic calibration procedure to be directly transferable to the robot controller, we used the DH convention to derive the kinematic model of the manipulator. In particular, we followed the approach described in \citep{Multiple-Single-Encoder} to derive the so-called DH(-) calibration model. As stated in Section \ref{SEC:introduction}, a kinematic calibration model should meet the three principles of completeness, continuity, and minimality. The standard DH model is not parametrically continuous for the Adept Viper S650 as joint axes 2 and 3 are parallel: a minor misalignment between those axes can result in major changes to the remaining DH parameters. For this reason, as in \citep{LaserTracker1,Ballbar1,Multiple-Single-Encoder}, we chose not to perturb the link twist $\alpha_2$. In addition, in our robot controller it was not possible to modify the nominally zero-valued DH parameters: consequently, we did not consider those parameters in our calibration model, as in \citep{Multiple-Single-Encoder}. Furthermore, using the proposed calibration approach, the error related to the first joint angle $\delta \theta_1$ cannot be estimated, as this quantity shifts all calibration points by the same amount, while the distances between these points (which are the measurements used) remain unchanged. Finally, to avoid redundacy, either the error parameters $\delta \theta_6$ and $\delta d_6$ or the error parameters related to the position of the tool frame ($\delta x_{tool}$, $\delta y_{tool}$, $\delta z_{tool}$) can be estimated. If the tool used during calibration is not the same as that used during robot operations, finding the error parameters related to the tool position is not effective. For this reason, we used a tool of known dimensions mounted directly on the robot and decided to estimate the errors associated with the last link, i.e. $\delta \theta_6$ and $\delta d_6$. The resulting kinematic calibration model is presented in Table \ref{TAB:calibration_DH} and is composed of only 10 error parameters, which can be grouped as
\begin{equation}
    \bm{\delta e} = \begin{bmatrix}
        \delta \theta_2 & \delta \theta_3 & \delta \theta_4 & \delta \theta_5 & \delta \theta_6 & \delta a_1 & \delta a_2 & \delta a_3 & \delta d_4 & \delta d_6
    \end{bmatrix}.
\end{equation}
Although this kinematic calibration model is incomplete, it has been proven to be robust and effective \citep{Multiple-Single-Encoder}.
\begin{table}[]
    \caption{DH kinematic calibration model for the Adept Viper S650 manipulator.}
    \label{TAB:calibration_DH}
    \centering
    \begin{tabular}{ccccc}
    \toprule
    $i$ & $\alpha_{i-1}$ & $a_{i-1}$ & $\theta_i$ & $d_i$ \\
    & $(\unit{\degree})$ & $(\unit{\milli\meter})$ & $(\unit{\degree})$ & $(\unit{\milli\meter})$ \\
    \midrule
    $1$ & $0$ & $0$ & $\theta_1$ & $0$ \\
    $2$ & $-90$ & $75+\delta a_1$ & $\theta_2+\delta \theta_2$ & $0$ \\
    $3$ & $0$ & $270+\delta a_2$ & $\theta_3+\delta \theta_3$ & $0$ \\
    $4$ & $90$ & $-90+\delta a_3$ & $\theta_4+\delta \theta_4$ & $295+\delta d_4$ \\
    $5$ & $-90$ & $0$ & $\theta_5+\delta \theta_5$ & $0$ \\
    $6$ & $90$ & $0$ & $\theta_6+\delta \theta_6$ & $80+ \delta d_6$ \\
    \bottomrule
    \end{tabular}
\end{table}

\subsection{Encoder location and calibration points}
\label{SUBSEC:choiceofpoints}
The proposed calibration approach first requires the definition of the joint coordinates $\bm{\theta_0}$ that identify the position of the draw-wire encoder. The draw-wire encoder should be placed in a position within the robot's workspace where it can be easily set-up and where it does not hinder the robot's movements, taking into account potential obstacles that may be present near the robot. For this reason, we chose $\bm{\theta_0}=[0,-90,210,-90,0,-90]$ as the initial configuration of the robot. Once $\bm{\theta_0}$ has been defined, the kinematic calibration procedure then requires the identification of a different set of calibration points for each unknown parameter. This is carried out by analyzing matrix $\bm{\Psi}_i$, which for our manipulator can be expressed as
\begin{equation}
    \bm{\Psi_i} = \begin{bmatrix}
        \Psi_{i_{\theta_2}} & \Psi_{i_{\theta_3}} & \Psi_{i_{\theta_4}} & \Psi_{i_{\theta_5}} & \Psi_{i_{\theta_6}} & \Psi_{i_{a_1}} & \Psi_{i_{a_2}} & \Psi_{i_{a_3}} & \Psi_{i_{d_4}} & \Psi_{i_{d_6}}
    \end{bmatrix},
\end{equation}
with different values of joint coordinates $\bm{\theta_i}$. This analysis also determines the order in which the error parameters are identified. Because our robot is an open chain manipulator, the first error parameters that can be estimated are the one closest to the end of the kinematic chain (i.e., the ones related to the last link) and subsequently all of the remaining ones. In fact, the analysis of $\bm{\Psi_i}$ with different values of $\bm{\theta_i}$ determined that the order in which the parameters are identified is: $\delta \theta_6$, $\delta d_6$, $\delta \theta_5$, $\delta \theta_4$, $\delta a_3$, $\delta d_4$, $\delta a_2$, $\delta \theta_3$, $\delta \theta_2$ and $\delta a_1$. Consequently, the error vector $\bm{\delta e}$ can be rearranged as 
\begin{equation}
    \bm{\overline{\delta e}} = \begin{bmatrix}
        \delta \theta_6 & \delta d_6 & \delta \theta_5 & \delta \theta_4 & \delta a_3 & \delta d_4 & \delta a_2 & \delta \theta_3 & \delta \theta_2 & \delta a_1
    \end{bmatrix},
\end{equation}
and matrix $\bm{\Psi_i}$ can be rearranged as
\begin{equation}
    \bm{\overline{\Psi}_i} = \begin{bmatrix}
        \Psi_{i_{\theta_6}} & \Psi_{ i_{d_6}} & \Psi_{i_{\theta_5}} & \Psi_{i_{\theta_4}} & \Psi_{i_{a_3}} & \Psi_{i_{ d_4}} & \Psi_{i_{a_2}} & \Psi_{i_{\theta_3}} & \Psi_{i_{ \theta_2}} & \Psi_{i_{a_1}}
    \end{bmatrix}.
\end{equation}
After determining the order in which the error parameters are identified, we defined the 10 sets of calibration points, one for each unknown error parameter. Because the analytical expressions for the calibration points that satisfy Equation (18) and ensure the highest value of $\Psi_{i_j}$ is quite difficult to obtain, we used an optimization program to find such points. The joint coordinates $\bm{\theta_i}$ and the corresponding matrix $\bm{\overline{\Psi}_i}$ of each calibration point are reported in Table \ref{TAB:calibpoints}. 


Before the experiments were performed, computer simulations were conducted to verify the correct functioning of the step-by-step kinematic calibration procedure and the goodness of the sets of calibration points. This was achieved by perturbing the nominal parameters by known amounts, producing artificial measured distance data (also taking into account the encoder's resolution and random noise), and checking that the identification process converged to the perturbed values. This step was necessary not only to validate the proposed approach but also because, for some parameters, we could not find points where $\Psi_{i_t}=0 \ ( \forall \ t \neq j)$ was exactly zero. For example, from Table \ref{TAB:calibpoints} it can be seen that the distance error at the calibration points used for the identification of $\delta \theta_4$ also depends slightly on $\delta a_3$ and $\delta d_4$ (both of which will be identified after $\delta \theta_4$). Similarly, the distance error at the calibration points used for the identification of $\delta a_3$ depends slightly on $\delta d_4$ (which is identified in the following step). Nonetheless, simulations carried out with different values of the error parameters showed that the identification process converged to the correct values.

\scriptsize
\setlength{\tabcolsep}{1.7pt}
\begin{longtable}{cccccccccccccccc}
\caption{Joint coordinates $\bm{\theta_i}$ and corresponding matrix $\bm{\overline{\Psi}_i}$ of the different calibration points}
\label{TAB:calibpoints} \\
\toprule
         $\theta_1$ & $\theta_2$ & $\theta_3$ & $\theta_4$ & $\theta_5$ & $\theta_6$ & $\Psi_{i_{\theta_6}}$ & $\Psi_{ i_{d_6}}$ & $\Psi_{i_{\theta_5}}$
         & $\Psi_{i_{\theta_4}}$ & $\Psi_{i_{a_3}}$ & $\Psi_{i_{ d_4}}$ & $\Psi_{i_{a_2}}$ & $\Psi_{i_{\theta_3}}$ & $\Psi_{i_{ \theta_2}}$ & $\Psi_{i_{a_1}}$ \\
         \midrule
\endfirsthead
\multicolumn{16}{c}{\tablename~\thetable\enspace(continued from previous page)}\\
\toprule
         $\theta_1$ & $\theta_2$ & $\theta_3$ & $\theta_4$ & $\theta_5$ & $\theta_6$ & $\Psi_{i_{\theta_6}}$ & $\Psi_{ i_{d_6}}$ & $\Psi_{i_{\theta_5}}$
         & $\Psi_{i_{\theta_4}}$ & $\Psi_{i_{a_3}}$ & $\Psi_{i_{ d_4}}$ & $\Psi_{i_{a_2}}$ & $\Psi_{i_{\theta_3}}$ & $\Psi_{i_{ \theta_2}}$ & $\Psi_{i_{a_1}}$ \\
         \midrule
\endhead
\midrule
\multicolumn{16}{r}{{Continued on next page}} \\ 
\endfoot
\endlastfoot
         \multicolumn{16}{l}{\textbf{Step 1: identification of $\bm{\delta \theta_6}$}} \\
         0 & -90 & 210 & -90 & -26 & -180 & -1.07 & 0 & 0 & 0 & 0 & 0 & 0 & 0 & 0 & 0  \\
         0 & -90 & 210 & -90 & -26 & -170 & -1.08 & 0 & 0 & 0 & 0 & 0& 0 & 0 & 0 & 0  \\
         0 & -90 & 210 & -90 & -26 & -190 & -1.07 & 0 & 0 & 0 &  0 & 0 & 0 & 0 & 0 & 0  \\
         0 & -90 & 210 & -90 & 26 & 0 & 1.07 & 0 & 0 & 0 & 0 & 0 & 0 & 0 & 0 & 0  \\
         0 & -90 & 210 & -90 & 26 & 10 & 1.07 & 0 & 0 & 0 & 0 & 0 & 0 & 0 & 0 & 0  \\
         0 & -90 & 210 & -90 & 26 & -10 & 1.08 & 0 & 0 & 0 & 0 & 0 & 0 & 0 & 0 & 0  \\
         \midrule
         \multicolumn{16}{l}{\textbf{Step 2: identification of $\bm{\delta d_6}$}} \\
         0 & -90 & 210 & -90 & -90 & -90 & 0 & 1.41 & 0 & 0 & 0 & 0 & 0 & 0 & 0 & 0  \\
         0 & -90 & 210 & -90 & -90 & -85 & 0.03 & 1.41 & 0 & 0 & 0 & 0 & 0 & 0 & 0 & 0  \\
         0 & -90 & 210 & -90 & -90 & -95 & -0.03 & 1.41 & 0 & 0 & 0 & 0 & 0 & 0 & 0 & 0  \\
         0 & -90 & 210 & -90 & 90 & -90 & 0 & 1.41 & 0 & 0 & 0 & 0 & 0 & 0 & 0 & 0  \\
         0 & -90 & 210 & -90 & 90 & -85 & 0.03 & 1.41 & 0 & 0 & 0 & 0 & 0 & 0 & 0 & 0  \\
         0 & -90 & 210 & -90 & 90 & -95 & -0.03 & 1.41 & 0 & 0 & 0 & 0 & 0 & 0 & 0 & 0  \\
         \midrule
         \multicolumn{16}{l}{\textbf{Step 3: identification of $\bm{\delta \theta_5}$}} \\
         0 & -90 & 210 & -135 & 10 & -90 & 0.09 & 0.17 & 0.2 & 0 & 0 & 0 & 0 & 0 & 0 & 0  \\
          0 & -90 & 210 & -135 & -5 & -90 & 0.1 & -0.08 & -0.22 & 0 & 0 & 0 & 0 & 0 & 0 & 0  \\
          0 & -90 & 210 & 45 & 0 & -270 & 0 & 0 & -4.19 & 0 & 0 & 0 & 0 & 0 & 0 & 0  \\
          0 & -90 & 210 & 45 & 10 & -270 & 0.29 & -0.13 & -3.9 & 0 & 0 & 0 & 0 & 0 & 0 & 0  \\
        0 & -90 & 210 & 45 & -5 & -270 & -0.06 & 0.08 & -4.17 & 0 & 0 & 0 & 0 & 0 & 0 & 0  \\
        0 & -90 & 210 & -45 & -5 & -90 & 0.06 & 0.08 & -0.13 & 0 & 0 & 0 & 0 & 0 & 0 & 0  \\
        0 & -90 & 210 & -45 & 10 & -90 & -0.29 & -0.13 & 0.62 & 0 & 0 & 0 & 0 & 0 & 0 & 0  \\
        0 & -90 & 210 & 135 & 0 & -270 & 0 & 0 & 4.19 & 0 & 0 & 0 & 0 & 0 & 0 & 0 \\
        0 & -90 & 210 & 135 & -5 & -270 & -0.1 & -0.08 & 4.15 & 0 & 0 & 0 & 0 & 0 & 0 & 0 \\
        0 & -90 & 210 & 135 & 10 & -270 & 0.09 & 0.17 & 4.13 & 0 & 0 & 0 & 0 & 0 & 0 & 0 \\
        \midrule
         \multicolumn{16}{l}{\textbf{Step 4: identification of $\bm{\delta \theta_4}$}} \\
        0 & -90 & 206 & 0 & 45 & -90 & -1.3 & 0.18 & 2.82 & 0.43 & -0.04 & 0 & 0 & 0 & 0 & 0 \\
        0 & -90 & 206 & 10 & 45 & -90 & -1.29 & 0.17 & 2.8 & 0.46 & -0.03 & 0.01 & 0 & 0 & 0 & 0 \\
        0 & -90 & 206 & -10 & 45 & -90 & -1.29 & 0.22 & 2.81 & 0.38 & -0.05 & -0.01 & 0 & 0 & 0 & 0 \\
        0 & -90 & $239.93$ & 0 & $-78.93$ & -90 & -0.46 & 0.18 & 0.99 & -3.04 & 0.27 & 0.08 & 0 & 0 & 0 & 0 \\
        0 & -90 & $239.93$ & 10 & $-78.93$ & -90 & -0.16 & 0.21 & 0.34 & -2.62 & 0.36 & 0.03 & 0 & 0 & 0 & 0 \\
        0 & -90 & $239.93$ & -10 & $-78.93$ & -90 & -0.58 & 0.22 & 1.26 & -3.15 & 0.22 & 0.13 & 0 & 0 & 0 & 0 \\
        0 & -90 & $239.93$ & 0 & $-78.93$ & 90 & 0.46 & 0.18 & -2.81 & 3.04 & 0.27 & 0.08 & 0 & 0 & 0 & 0 \\
        0 & -90 & $239.93$ & 10 & $-78.93$ & 90 & 0.58 & 0.22 & -2.67 & 3.15 & 0.22 & 0.13 & 0 & 0 & 0 & 0 \\
        0 & -90 & $239.93$ & -10 & $-78.93$ & 90 & 0.16 & 0.21 & -2.93 & 2.26 & 0.36 & 0.03 & 0 & 0 & 0 & 0 \\
        0 & -90 & 206 & 0 & 45 & 90 & 0.13 & 0.18 & -0.97 & -0.43 & -0.04 & 0 & 0 & 0 & 0 & 0 \\
        0 & -90 & 206 & 10 & 45 & 90 & 0.29 & 0.22 & -0.49 & -0.38 & -0.05 & -0.01 & 0 & 0 & 0 & 0 \\
        0 & -90 & 206 & -10 & 45 & 90 & 0.29 & 0.17 & -1.19 & -0.46 & -0.03 & 0.01 & 0 & 0 & 0 & 0 \\
        \midrule
        \multicolumn{16}{l}{\textbf{Step 5: identification of $\bm{\delta a_3}$}} \\
        0 & -90 & 230 & -65 & -80 & 0 & -1.15 & 1.15 & 0.19 & -1.72 & 0.25 & 0 & 0 & 0 & 0 & 0 \\
        0 & -90 & 230 & -60 & -80 & 0 & -1.13 & 1.13 & 0.1 & -1.69 & 0.26 & -0.02 & 0 & 0 & 0 & 0 \\
        0 & -90 & 230 & -70 & -80 & 0 & -1.16 & 1.17 & 0.26 & -1.73 & 0.25 & 0.02 & 0 & 0 & 0 & 0 \\
        0 & -90 & 230 & 65 & -80 & 0 & 1.15 & 1.15 & -2.92 & 1.72 & 0.25 & 0 & 0 & 0 & 0 & 0 \\ 
        0 & -90 & 230 & 70 & -80 & 0 & 1.16 & 1.17 & -2.88 & 1.73 & 0.25 & 0.02 & 0 & 0 & 0 & 0 \\
        0 & -90 & 230 & 60 & -80 & 0 & 1.13 & 1.13 & -2.96 & 1.69 & 0.26 & -0.02 & 0 & 0 & 0 & 0 \\
        \midrule
        \multicolumn{16}{l}{\textbf{Step 6: identification of $\bm{\delta d_4}$}} \\
        0 & -90 & 130 & -90 & 0 & -90 & 0 & 1.28 & 0 & 0 & -0.09 & 1.28 & 0 & 0 & 0 & 0 \\
        0 & -90 & 135 & -90 & 0 & -90 & 0 & 1.21 & 0 & 0 & -0.09 & 1.21 & 0 & 0 & 0 & 0 \\
        0 & -90 & 140 & -90 & 0 & -90 & 0 & 1.14 & 0 & 0 & -0.08 & 1.14 & 0 & 0 & 0 & 0 \\
        0 & -90 & 125 & -90 & 0 & -90 & 0 & 1.35 & 0 & 0 & -0.1 & 1.35 & 0 & 0 & 0 & 0 \\
        0 & -90 & 120 & -90 & 0 & -90 & 0 & 1.41 & 0 & 0 & -0.1 & 1.41 & 0 & 0 & 0 & 0 \\
        0 & -90 & 150 & -90 & 0 & -90 & 0 & 1.0 & 0 & 0 & -0.07 & 1.0 & 0 & 0 & 0 & 0 \\
        \midrule
        \multicolumn{16}{l}{\textbf{Step 7: identification of $\bm{\delta a_2}$}} \\
        0 & -110 & 230 & -90 & 0 & -90 & 0 & 0 & 0 & 0 & 0 & 0 & 0.35 & 0 & 0 & 0 \\
        0 & -110 & 235 & -90 & 0 & -90 & 0 & 0 & 0 & 0 & 0 & 0 & 0.43 & 0 & 0 & 0 \\
        0 & -110 & 240 & -90 & 0 & -90 & 0 & 0 & 0 & 0 & 0 & 0 & 0.52 & 0 & 0 & 0 \\
        0 & -110 & 245 & -90 & 0 & -90 & 0 & 0 & 0 & 0 & 0 & 0 & 0.6 & 0 & 0 & 0 \\
        0 & -110 & 250 & -90 & 0 & -90 & 0 & 0 & 0 & 0 & 0 & 0 & 0.68 & 0 & 0 & 0 \\
        0 & -110 & 255 & -90 & 0 & -90 & 0 & 0 & 0 & 0 & 0 & 0 & 0.77 & 0 & 0 & 0 \\
        \midrule
        \multicolumn{16}{l}{\textbf{Step 8: identification of $\bm{\delta \theta_3}$}} \\
        0 & -110 & 175 & -90 & 0 & -90 & 0 & 0.88 & 0 & 0 & -0.28 & 0.88 & 0.03 & -1.63 & 0 & 0 \\  
        0 & -110 & 180 & -90 & 0 & -90 & 0 & 0.8 & 0 & 0 & -0.28 & 0.8 & 0.02 & -1.63 & 0 & 0 \\ 
        0 & -110 & 185 & -90 & 0 & -90 & 0 & 0.72 & 0 & 0 & -0.27 & 0.72 & 0.02 & -1.63 & 0 & 0 \\ 
        0 & -120 & 190 & -90 & 0 & -90 & 0 & 0.76 & 0 & 0 & -0.38 & 0.76 & 0.06 & -2.42 & 0 & 0 \\ 
        0 & -120 & 195 & -90 & 0 & -90 & 0 & 0.67 & 0 & 0 & -0.37 & 0.67 & 0.06 & -2.42 & 0 & 0 \\
        0 & -120 & 185 & -90 & 0 & -90 & 0 & 0.84 & 0 & 0 & -0.39 & 0.84 & 0.06 & -2.42 & 0 & 0 \\
        \midrule
        \multicolumn{16}{l}{\textbf{Step 9: identification of $\bm{\delta \theta_2}$}} \\
        -20 & -95 & 210 & -90 & -90 & 0 & 0.39 & 0.8 & -1.93 & 0.95 & -0.08 & 0.04 & 0.05 & 0.02 & 0.39 & 0.01 \\
        -20 & -90 & 210 & -90 & -90 & 0 & 0.36 & 0.77 & -2.06 & 0.85 & -0.01 & 0.01 & 0 & 0.34 & 0.41 & 0.01 \\
        -20 & -85 & 210 & -90 & -90 & 0 & 0.32 & 0.74 & -2.21 & 0.72 & 0.07 & -0.01 & -0.05 & 0.63 & 0.42 & 0.02 \\
        -20 & 0 & 15 & -90 & -90 & 0 & 0.71 & 0.77 & 1.49 & 1.62 & -0.31 & 1.55 & -1.38 & 1.65 & 0.41 & 0.01 \\
        -20 & -5 & 15 & -90 & -90 & 0 & 0.65 & 0.79 & 1.66 & 1.52 & -0.3 & 1.6 & -1.31 & 1.67 & 0.4 & 0 \\
        -20 & 5 & 15 & -90 & -90 & 0 & 0.78 & 0.74 & 1.29 & 1.73 & -0.33 & 1.5 & -1.43 & 1.58 & 0.43 & 0.02 \\
        \midrule
        \multicolumn{16}{l}{\textbf{Step 10: identification of $\bm{\delta a_1}$}} \\
        -160 & -125 & 5 & -90 & -90 & 0 & 0.97 & 0.54 & -2.14 & 1.85 & -1.44 & -0.01 & -0.49 & -8.99 & -1.68 & 1.89 \\
        -160 & -120 & 5 & -90 & -90 & 0 & 0.8 & 0.63 & -2.48 & 1.47 & -1.17 & -0.06 & -0.47 & -7.72 & 0.46 & 1.94 \\
        -160 & -115 & 5 & -90 & -90 & 0 & 0.6 & 0.71 & -2.73 & 1.03 & -0.86 & -0.09 & -0.42 & -6.22 & 2.58 & 1.97 \\
        -160 & -110 & 5 & -90 & -90 & 0 & 0.39 & 0.77 & -2.85 & 0.58 & -0.55 & -0.09 & -0.34 & -4.62 & 4.49 & 1.96 \\
        -160 & -105 & 5 & -90 & -90 & 0 & 0.19 & 0.81 & -2.88 & 0.14 & -0.25 & -0.06 & -0.26 & -3.08 & 6.08 & 1.92 \\
         \bottomrule
\end{longtable}
\normalsize
\setlength{\tabcolsep}{6pt}

\subsection{Experimental setup of the calibration procedure}
\label{SUBSEC:calibrationprocedure}
Given the effects of temperature changes on the geometrical properties of the manipulator, all test were performed while monitoring both the room temperature and the temperature of the motor driving circuits. In fact, experimental tests only started when the motor amplifier temperature stabilized to its steady state value after the initial transient that starts when enabling robot power. More in detail, the room temperature was kept constant at $\SI{25}{\degree}$, while the temperature of the motor driving circuits stabilized between $\SI{57}{\degree}$ and $\SI{59}{\degree}$. The kinematic calibration of the Adept Viper S650 was then carried out following the proposed step-by-step procedure (described in the flowchart in Figure \ref{FIG:flowchart}), moving the robot through the sets of calibration points reported in Table \ref{TAB:calibpoints}. More in detail, at the beginning of each step of the procedure, the robot was moved to the configuration specified by the joint coordinates $\bm{\theta_0}$: at each step we checked whether the measured distance was still zero, to ensure that the encoder was working properly and no wire deformation occurred. The robot was then moved to the set of calibration points; at each calibration point, we waited three seconds before recording the wire length, to allow any dynamic oscillations to subside. In addition, moving the robot between a set of calibration points without twisting the wire required the definition of a number of intermediate path points (via points). Finally, once the calibration procedure ended and we estimated all error parameters, we implemented error compensation by modifying the nominal parameters embedded in the robot controller. 

\begin{figure}
    \centering
    \includegraphics[width=0.9\textwidth]{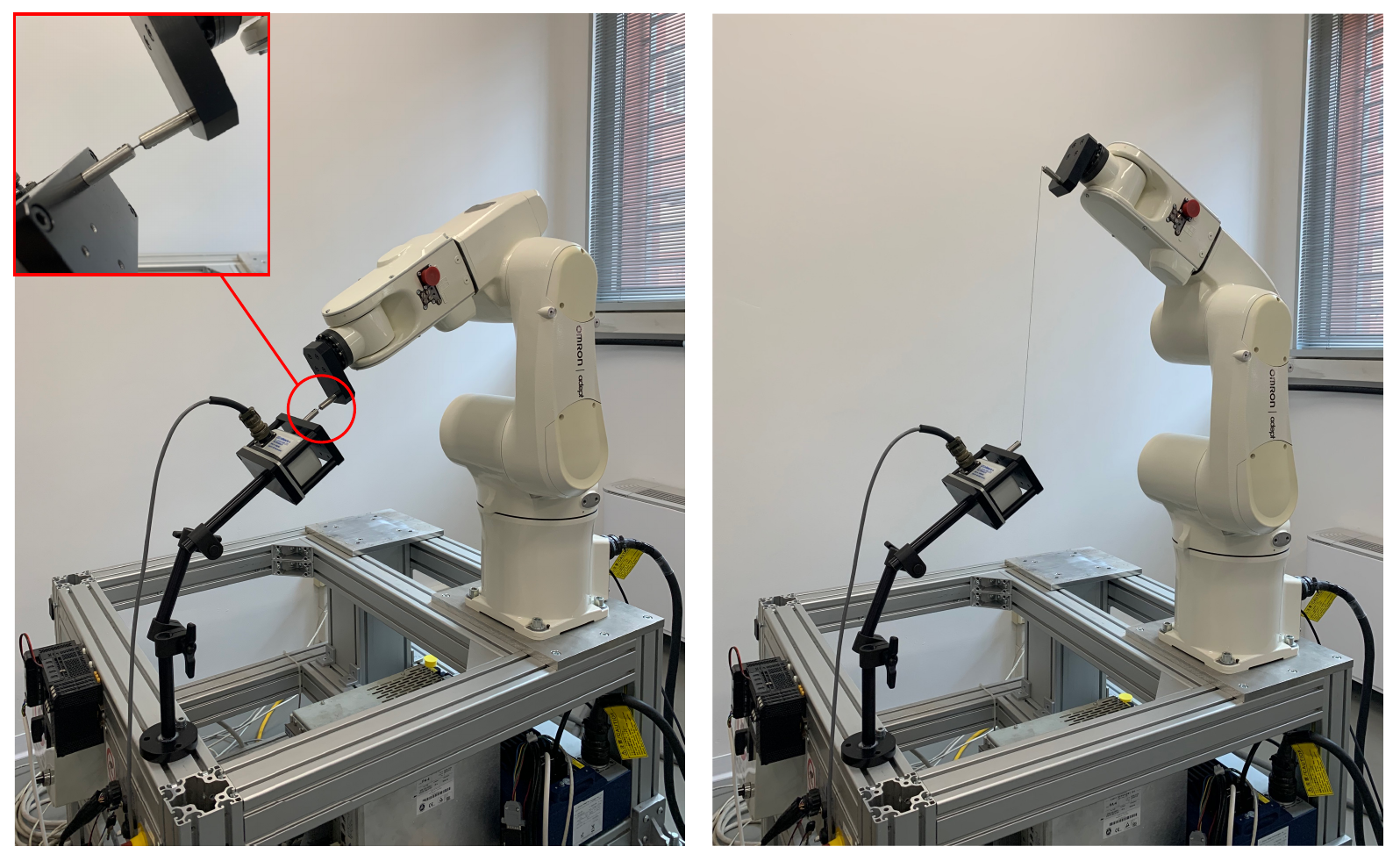}
    \caption{Experimental setup during the calibration procedure: system at the configuration specified by the joint coordinates $\theta_0$ (left) and at the $i$th calibration configuration, specified by the joint coordinates $\theta_i$ (right).}
    \label{FIG:calibrationprocedure}
\end{figure}

Figure \ref{FIG:calibrationprocedure} shows the manipulator during the calibration procedure in the configuration described by $\bm{\theta_0}$ (where the encoder is mounted so that the measured distance is zero) and in a generic configuration $\bm{\theta_i}$. 

\subsection{Validation procedure}
Validation is often carried out using the same end-effector and measurement system used to acquire measurement data for calibration: this could produce biased results and an overestimation of the resulting accuracy after calibration. For this reason, we carried out the validation of our kinematic calibration procedure following two different approaches: the first uses the draw-wire encoder, while the second uses a touch probe and a granite surface plate.
\subsubsection{Validation procedure using the draw-wire encoder}
\label{SUBSEC:validationprocedure1}
When moving a robot within its workspace, the target pose can be specified in the Cartesian space (i.e., with respect to a reference frame) or in the joint space (i.e., specifying each joint variable). If a pose in the robot's workspace is specified in the Cartesian space, the controller of the robot computes the joint angles necessary to reach that pose through the inverse kinematics. For a specific pose in the robot's workspace, there could be multiple solutions to the inverse kinematics and they are usually referred to as robot configurations \citep{Craig}. For a 6 DOF anthropomorphic arm with a spherical wrist, the same pose can be reached with up to 8 different configurations. In particular, there are two possible solutions for $\theta_1$ (typically referred to as "righty" or "lefty"), two possible solutions for $\theta_3$ ("elbow above" or "elbow below") and two possible solutions for $\theta_4$ ("nonflip" or "flip"). Given a well-calibrated robot, if the configuration of the manipulator is changed while the commanded pose remains the same, the Cartesian coordinates should remain unchanged, as well as the distance between the end-effector and a fixed location. Therefore, the variation of this distance can be measured with the draw encoder and used as a metric to evaluate the robot's accuracy. 

To assess the robot's accuracy before and after calibration, we fist placed the draw-wire encoder outside of the manipulator's workspace, in order not to hinder the robot's movements. We then moved the robot through a set of 50 different poses within the entire robot's workspace, and at each pose the robot's configuration was changed between "elbow above"/"elbow below" and "nonflip/flip" (four possible combinations). At each robot pose and configuration, the distance measured by the draw wire encoder was recorded. The distances corresponding to the same end-effector pose were compared, and the maximum discrepancy between them was used as a performance metric. 
\begin{figure}
    \centering
    \includegraphics[width=0.9\textwidth]{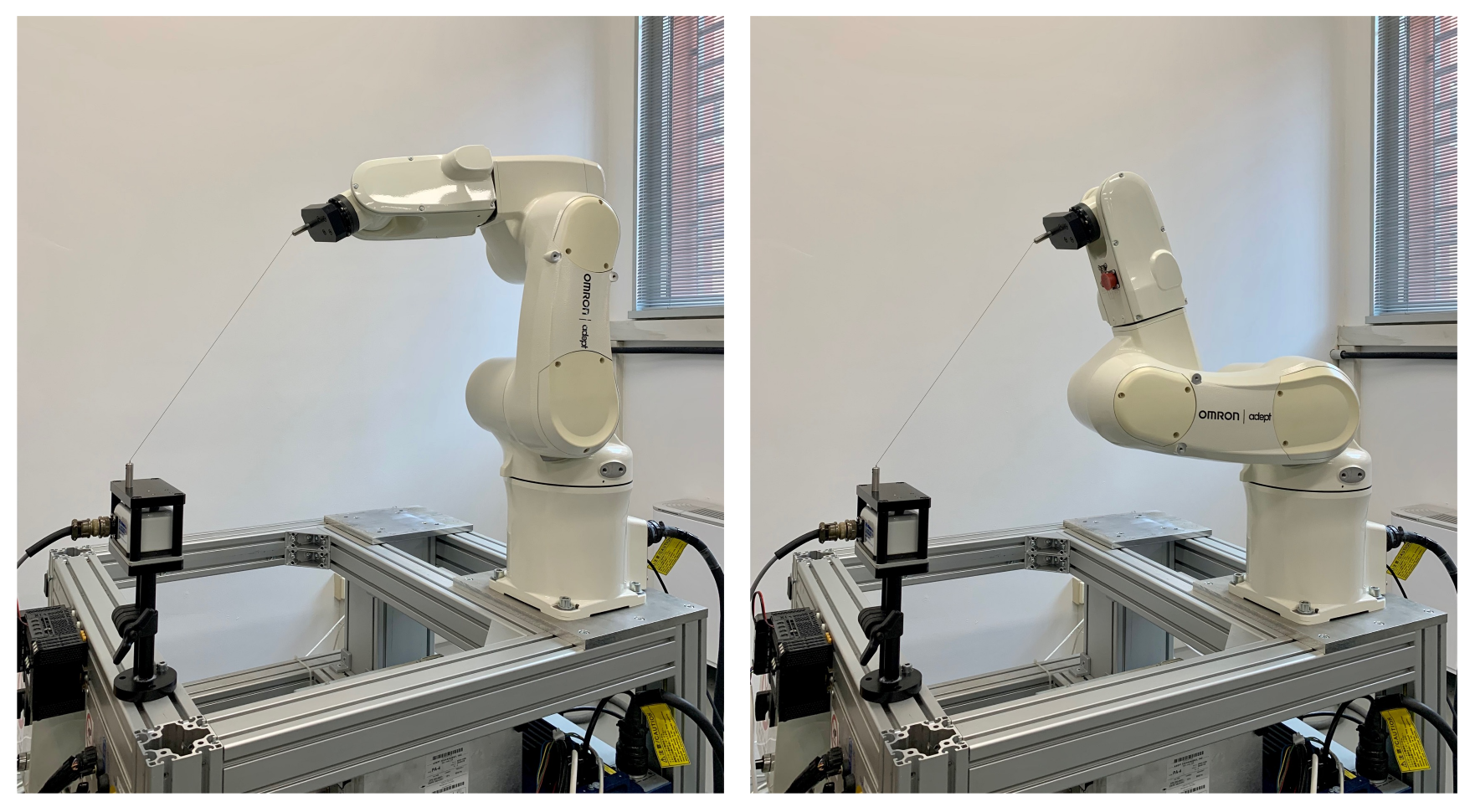}
    \caption{Experimental setup during the first validation procedure, where the robot is moved through a set of poses in the workspace, changing its configuration: "above" and "noflip" (left), "below" and "flip" (right) }
    \label{FIG:validationencoder}
\end{figure}

\subsubsection{Validation procedure using a touch probe and a granite surface plate}
\label{SUBSEC:validationprocedure2}

To evaluate the accuracy of the robot before and after calibration, we also used a touch probe mounted on the robot's flange and a granite surface plate placed on the table on which the robot stands. The faces of a granite surface plate are machined with a strict flatness tolerance: all points on the faces lie on the same plane following the planar equation
\begin{equation}
    z_i=ax_i+by_i+c,
    \label{EQ:plane}
\end{equation}
where $a$, $b$ and $c$ are the constant coefficients that characterize the plane. By using a touch probe, we can move the robot's end-effector until it touches the plate and record the corresponding end-effector coordinates computed by the robot controller. When a suitable number of data points have been acquired, we can perform the best surface fit in the least-squares sense. The fit of the plane (i.e. the values of the residuals) reflects the accuracy of the robot itself. In fact, this quantity has been effectively used as a cost function for robot calibration in \citep{PSO1}; here, we only use it to assess the robot's accuracy before and after calibration. For this purpose, the recorded coordinates of acquisition points can be grouped in the following column vectors
\begin{equation}
    \bm{X}=\begin{bmatrix}
        x_1 \\
        \vdots \\
        x_n
    \end{bmatrix} \quad 
    \bm{Y}=\begin{bmatrix}
        y_1 \\
        \vdots \\
        y_n
    \end{bmatrix} \quad \bm{Z}=\begin{bmatrix}
        z_1 \\
        \vdots \\
        z_n
    \end{bmatrix},
\end{equation}
and Equation \eqref{EQ:plane} can be rewritten in matrix form as
\begin{equation}
    \bm{Z}=a\bm{X}+b\bm{Y}+c .
\end{equation}
This is an overconstrained linear system: the coefficients $a$, $b$ and $c$ can be calculated by least linear squares using Moore-Penrose pseudoinverse. Let us first define 
\begin{equation}
    \bm{\Upsilon}= \begin{bmatrix}
        \bm{X} & \bm{Y} & \bm{1}
    \end{bmatrix},
\end{equation}
where $\bm{1}$ is an all-ones column vector of size $n$. Then, we can approximately solve for the coefficients as
\begin{equation}
    \begin{bmatrix}
        a \\
        b \\
        c 
    \end{bmatrix} = (\bm{\Upsilon}^T\bm{\Upsilon})^{-1}\bm{\Upsilon}^T \bm{Z}.
\end{equation}
For each data acquisition point, the residual from fitting the plane can be calculated as
\begin{equation}
    \Delta_i = ax_i+by_i+c-z_i,
\end{equation}
and this quantity can be used as a performance metric to evaluate the robot's accuracy before and after calibration.

\begin{figure}
    \centering
    \includegraphics[width=0.4\textwidth]{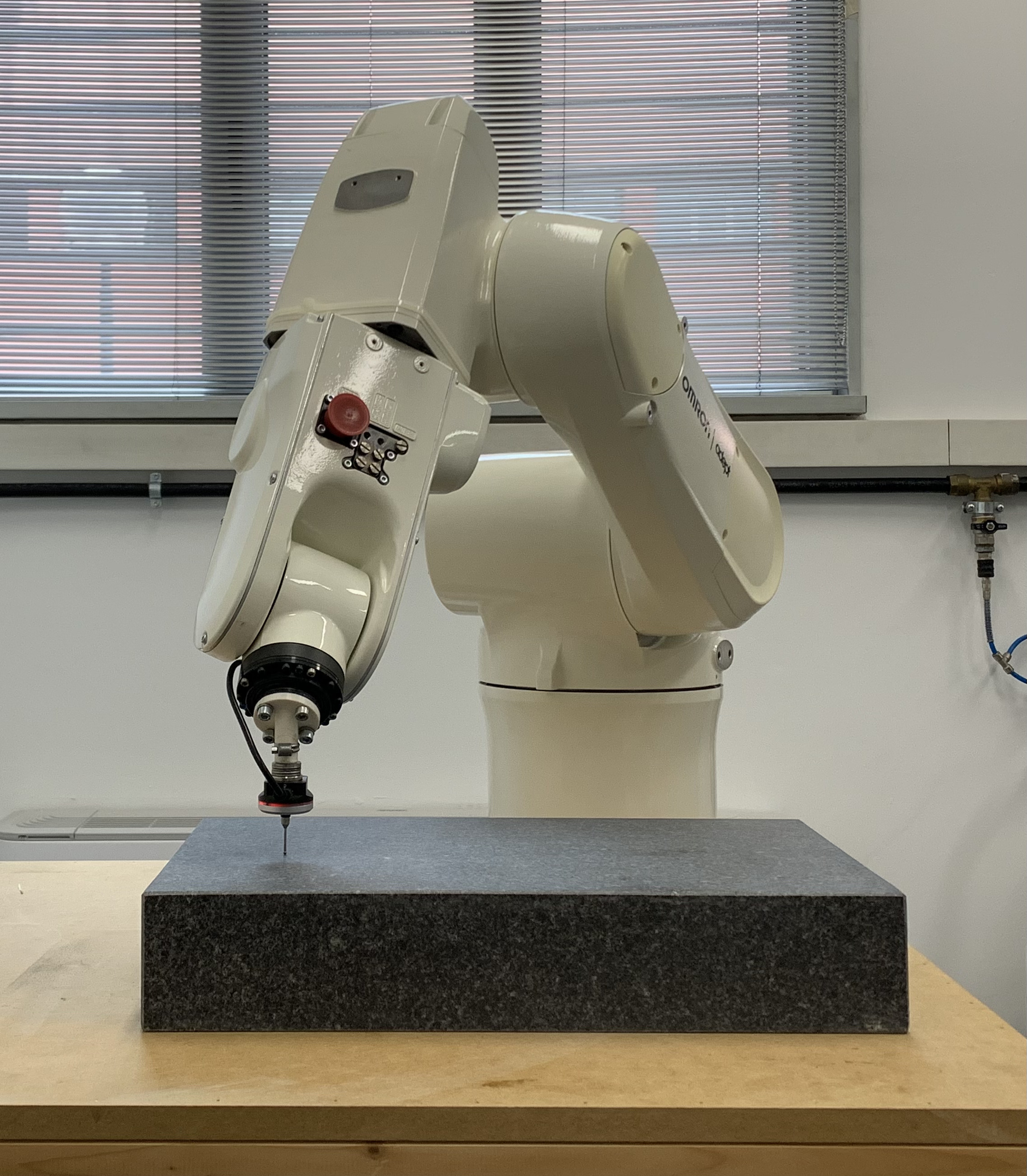}
    \caption{Experimental setup during the second validation procedure: the touch probe detects the contact with the granite surface plate (the LED light turns red) and the corresponding end-effector coordinates are recorded.}
    \label{FIG:validationtouchprobe}
\end{figure}

To carry out this validation procedure, we used a 3D touch probe directly connected to the robot controller; its specifications are a resolution of $\SI{1}{\micro\meter}$ and a maximum error of $\SI{4}{\micro\meter}$. The granite surface plate that we used is machined with grade 0 accuracy according to the DIN 876 standard and has a flatness tolerance of $\SI{6}{\micro\meter}$; its dimensions are $\qtyproduct{400 x 250 x 70}{\milli\meter}$. To acquire the end-effector coordinates of points that lie on the surface plate, we defined a $19 \times 11$ grid, for a total of $209$ acquisition points. The robot was first moved above each acquisition point and then moved along the $z$ direction until the touch probe detected a contact and the end-effector coordinates were recorded.

\section{Results} 
\label{SEC:results}
\subsection{Kinematic calibration results}
The kinematic calibration procedure described in Section \ref{SUBSEC:calibrationprocedure} was fully automated and took approximately 45 minutes to perform. The estimated error parameters are reported in Table \ref{TAB:calibrationresults}. The identified joint offset values appear significantly higher than the geometric parameter offset values: this suggests that most of the robot's inaccuracy before calibration is due to incorrect offset values used to describe the manipulator's home position.
\begin{table}[]
    \centering
    \caption{Kinematic calibration results.}
    \label{TAB:calibrationresults}
    \begin{tabular}{ccc}
    \toprule
         Parameter & Nominal value & Estimated offset value \\
        \midrule
        $\theta_2$ (\unit{\degree}) & $0.000$ & $0.675$ \\
        $\theta_3$ (\unit{\degree}) & $0.000$ & $-0.485$ \\
        $\theta_4$ (\unit{\degree}) & $0.000$ & $0.245$ \\
        $\theta_5$ (\unit{\degree}) & $0.000$ & $-0.575$ \\
        $\theta_6$ (\unit{\degree}) & $0.000$ & $-1.215$ \\
        $a_1$ (\unit{\milli\meter}) & $75.000$ & $-0.005$ \\
        $a_2$ (\unit{\milli\meter}) & $270.000$ & $0.105$ \\
        $a_3$ (\unit{\milli\meter}) & $-90.000$ & $0.025$ \\
        $d_4$ (\unit{\milli\meter}) & $295.000$ & $-0.105$ \\
        $d_6$ (\unit{\milli\meter}) & $80.000$ & $0.115$ \\
        \bottomrule
    \end{tabular}
\end{table}
\subsection{Validation results}
\subsubsection{Validation using the draw-wire encoder}
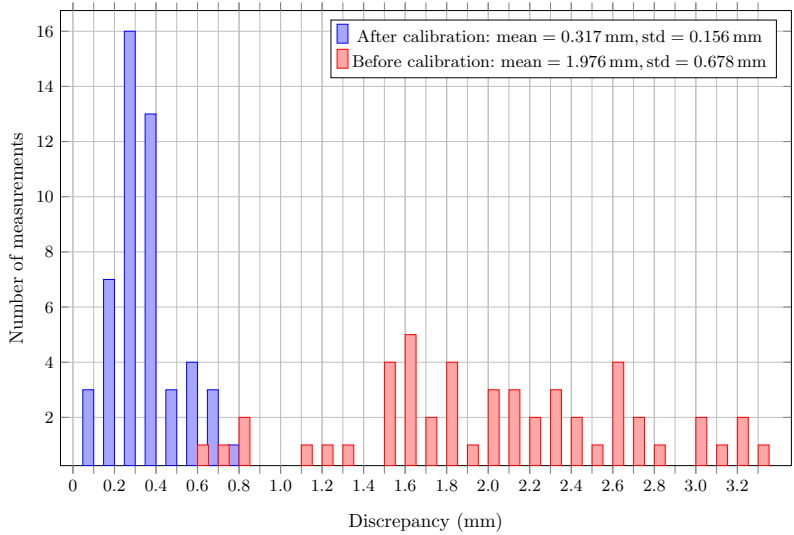
\begin{figure}
\centering
\resizebox{0.8\textwidth}{!}{%
\begin{tikzpicture}[scale=1]
\begin{axis} [
ybar=0cm,
height=10cm,
width=15cm,
bar width=0.2cm,
ticklabel style = {font=\small},
legend style = {font=\small},
xtick={0,0.1,0.2,...,3.3}, 
xticklabels={},
extra x ticks={0,0.2,0.4,0.6,0.8,1.0,1.2,1.4,1.6,1.8,2.0,2.2,2.4,2.6,2.8,3.0,3.2}, 
extra x tick labels={0,0.2,0.4,0.6,0.8,1.0,1.2,1.4,1.6,1.8,2.0,2.2,2.4,2.6,2.8,3.0,3.2},
enlargelimits=0.05, 
ylabel=Number of measurements,
xlabel=Discrepancy $\left(\unit{\milli\meter}\right)$,
grid=major,
legend image code/.code={
        \draw [#1] (0cm,-0.1cm) rectangle (0.2cm,0.25cm); },
]
\addplot[blue,fill=blue!35!white]
	coordinates {(0.1,3) (0.2,7) (0.3,16) (0.4,13) (0.5,3) (0.6,4) (0.7,3) (0.8,1)};
\addplot[red,fill=red!35!white]
    coordinates {(0.6,1) (0.7,1) (0.8,2) (1.1,1) (1.2,1) (1.3,1) (1.5,4) (1.6,5) (1.7,2) (1.8,4) (1.9,1) (2,3) (2.1,3) (2.2,2) (2.3,3) (2.4,2) (2.5,1) (2.6,4) (2.7,2) (2.8,1) (3,2) (3.1,1) (3.2,2) (3.3,1)};
\legend{After calibration: $\text{mean}=\SI{0.317}{\milli\meter},\text{std}=\SI{0.156}{\milli\meter}$ \\ Before calibration: $\text{mean}=\SI{1,976}{\milli\meter},\text{std}=\SI{0.678}{\milli\meter}$ \\}
\end{axis}
\end{tikzpicture} 
}
\caption{Maximum discrepancy among the recorded distances at the same end-effector pose but different configurations, observed in 50 different poses within the robot's workspace, before and after calibration.}
\label{FIG:encodervalidation}
\end{figure}

The validation procedure described in Section \ref{SUBSEC:validationprocedure1} was carried out before and after calibration: its results are reported in Figure \ref{FIG:encodervalidation} in the form of a histogram. Before calibration, the discrepancy among the recorded distances (at the same end-effector pose but different configuration) assumes values in the order of some millimeters and presents a high variability between measurements taken in different poses within the robot's workspace. After calibration, the discrepancy assumes values in the order of tenths of millimeters and the variation between different measures is rather small. In fact, after calibration, the mean value of the maximum discrepancy is reduced by 84\% while the standard deviation is reduced by 77\%.

\subsubsection{Validation using the touch probe and granite surface plate}
\begin{figure}
\centering
\resizebox{0.8\textwidth}{!}{%
\begin{tikzpicture}[scale=1]
\begin{axis} [
ybar=0cm,
ymin=0,
height=10cm,
ticklabel style = {font=\small},
legend style = {font=\small},
width=15cm,
bar width=0.195cm,
xtick={0,0.01,0.02,...,0.33}, 
xticklabels={},
extra x ticks={0,0.02,0.04,0.06,0.08,0.10,0.12,0.14,0.16,0.18,0.20,0.22,0.24,0.26,0.28,0.30,0.32}, 
extra x tick labels={0,0.02,0.04,0.06,0.08,0.10,0.12,0.14,0.16,0.18,0.20,0.22,0.24,0.26,0.28,0.30,0.32},
enlarge x limits=0.05,
ylabel=Number of measurements,
xlabel=Residuals $\left(\unit{\milli\meter}\right)$,
grid=major,
legend image code/.code={
        \draw [#1] (0cm,-0.1cm) rectangle (0.2cm,0.25cm); },
]
\addplot[blue,fill=blue!35!white]
	coordinates {(0.01,80) (0.02,69) (0.03,36) (0.04,20) (0.05,4)};
\addplot[red,fill=red!35!white]
    coordinates {(0.01,25) (0.02,24) (0.03,18) (0.04,19) (0.05,19) (0.06,9) (0.07,13) (0.08,10) (0.09,17) (0.1,9) (0.11,9) (0.12,8) (0.13,3) 
    (0.14,7) (0.15,4) (0.16,4) (0.17,1) 
    (0.18,2) (0.19,1) (0.21,1) (0.23,1) (0.24,3) (0.31,1) (0.32,1)
    };
\legend{After calibration: $\text{mean}=\SI{0.0148}{\milli\meter},\text{std}=\SI{0.0108}{\milli\meter}$ \\ 
Before calibration: $\text{mean}=\SI{0.0655}{\milli\meter},\text{std}=\SI{0.0563}{\milli\meter}$ \\
}
\end{axis}
\end{tikzpicture} 
}
\caption{Residuals of the plane fit given a set of $209$ points that lies on granite surface plate, before and after calibration.}
\label{FIG:touchprobe}
\end{figure}
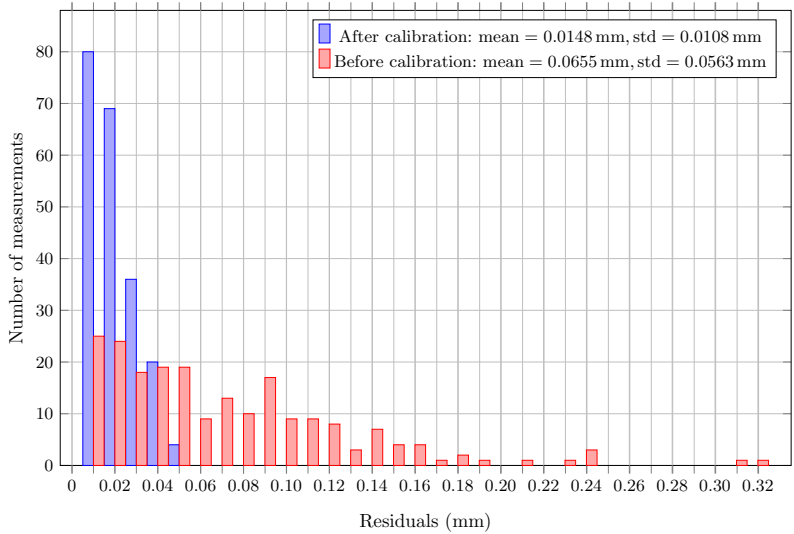

The second validation procedure, described in Section \ref{SUBSEC:validationprocedure2}, was carried out before and after calibration: its results are presented in Figure \ref{FIG:touchprobe} in the form of a histogram. Figure \ref{FIG:touchprobe} shows that the residuals of the plane fit are greatly reduced after calibration: before calibration, the residuals are below $\SI{0.32}{\milli\meter}$, while the maximum deviation obtained after calibration is less than $\SI{0.05}{\milli\meter}$. In particular, after calibration the mean value of the residuals is decreased by $77\%$ while the standard deviation is decreased by $80\%$.

\section{Discussion}
\label{SEC:discussion}

The results of the two different validation procedures showed a clear improvement in the robot accuracy after calibration, proving the effectiveness of the proposed step-by-step calibration procedure. More in detail, after calibration, the maximum discrepancy among the recorded distances in the same end-effector pose but with different configurations decreased from $\SI{3.3}{\milli\meter}$ to $\SI{0.8}{\milli\meter}$. A similar performance metric was used in \citep{Multiple-Single-Encoder}, where distances corresponding to the same end-effector pose but different orientations were recorded and compared: if this deviation is below $\SI{2}{\milli\meter}$, it is safe to assume that the robot is fairly well calibrated for most applications \citep{Multiple-Single-Encoder}. Since changing the robot configuration while keeping its end-effector pose constant requires a greater movement of the joints than changing its orientation, we can assume that our robot is well calibrated (while before calibration it was not). On the other hand, we cannot compare the results of the second validation procedure with other works, as in \citep{PSO1} the plane fit was used as a cost function during calibration, but validation was carried out using a CMM.  

The proposed calibration approach offers several advantages over standard calibration approaches. First, measurement data for the calibration of robots are typically obtained through laser trackers because of their high accuracy. However, they are very expensive, require special training to use, and their calibration is time consuming. Using a single draw-wire encoder, which is much less expensive than laser trackers, reduces the cost of calibration while effectively improving the robot's positioning accuracy. However, since we did not repeat the calibration procedure using laser trackers, it is impossible to claim with certainty that our procedure produces the same (or better) resulting accuracy after calibration. In addition, using standard approaches and measurement systems, the coordinates of the robot's end-effector are acquired in the coordinate system of the instrument and have to be transformed into the base coordinate system of the robot. This transformation typically requires the estimation of six additional parameters, increasing the computational burden of the identification process and potentially introducing errors that could propagate to the final results. Conversely, our approach does not require the explicit knowledge of a reference frame and identification of additional parameters, making calibration more robust. 

Our work also differs significantly from previous studies that used a single draw-wire encoder \citep{Multiple-Single-Encoder,Single-Encoder-1,Single-Encoder-2,Single-Encoder-3,Single-Encoder-4,Single-Encoder-5}. By addressing two aspects that were not addressed in previous works - the optimal encoder location and set of calibration points - we found that it is possible to define a set of calibration points for each unknown parameter: in these points, the difference between measured and predicted distance depends only on that parameter. As a result, we proposed a step-by-step calibration procedure where the parameters are subsequently identified, one at a time: each parameter is estimated faster and more accurately than with standard approaches. Following standard robot calibration procedures, the robot is moved to a number of calibration points where the measured error depends on all of the unknown error parameters; the error parameters are then estimated simultaneously and often with complex optimization algorithms (which cannot be implemented directly into the robot controllers due to their limited computational power). However, the optimization algorithm may converge to a set of error values that minimize the cost function but do not represent the actual error parameter values. Following the proposed procedure, only one of the unknown parameters is estimated at a time, and fewer measurements are processed at the same time: the identification process is more robust and, since it does not require complex optimization algorithms, it can be implemented directly into the robot controller, without the need for extra hardware to process the data. This is particularly advantageous in industrial environments where computing platforms such as Matlab may not be available.  

Nevertheless, the proposed calibration procedure has some potential limitations. The analytical approach used to find the sets of calibration points is based on the linearization of direct kinematics, which is effective only if the errors between actual and nominal parameters are relatively small. However, this may not be the case when calibration is performed after mechanical parts or batteries are replaced: in these instances a coarse calibration should be performed first. In addition, our calibration approach does not allow to find the error related to the first joint angle. 

Future studies should aim to compare the resulting accuracy after calibration obtained by the proposed step-by-step procedure with that obtained using standard optimization algorithms and more expensive measurement systems, such as laser trackers. In addition, a higher quality displacement sensor should be considered, in order to evaluate the peak performance of the presented calibration procedure. 

\section{Conclusions}
\label{SEC:conclusions}
In this paper, we presented a cost-effective and practical step-by-step kinematic calibration procedure for industrial robots using 1D measurement data obtained through a single draw-wire encoder. The heart of our approach lies in the proper choice of encoder location and calibration points. In particular, we positioned the draw-wire encoder in a location specified in the joint space and developed a novel analytical approach to find different sets of calibration points where the distance error depends only on one of the unknown error parameters. As a result, we proposed a step-by-step calibration procedure through which each error parameter is subsequently identified: this improves the identification accuracy while reducing the computational burden of the identification process. In fact, since our calibration procedure does not require complex optimization algorithms, it can be implemented directly into the robot controllers without the need for extra hardware to process measurement data. Numerical simulations and calibration experiments on a 6 DOF anthropomorphic arm proved the effectiveness and robustness of the proposed method, which represents a cost-effective and computationally less demanding alternative to standard calibration approaches.

\backmatter





\bmhead{Acknowledgments}

The authors would like to show their gratitude to Mr. Alessandro D'Amico who provided insight, materials, and expertise that greatly assisted the research.

\section*{Statements and Declarations}

\bmhead{Funding}

The authors declare that no funds, grants, or other support were received during the preparation of this manuscript.

\bmhead{Conflict of interest} 
The authors declare that they have no conflict of interest.

\begin{appendices}

\section{Proof of the proposed analytical approach}
\label{APP:1}

\begin{figure}
    \centering
    \resizebox{0.45\textwidth}{!}{%
\begin{tikzpicture}
\pic (MyBaseA) {frame pivot rounded=0.9cm};
\pic (L1) at (MyBaseA-center) {link bar generic=-25:60pt/0};
\pic (L2) at (L1-end) {link bar generic=35:90pt/0} ;
\pic (L3) at (MyBaseA-center) {link bar generic=45:60pt/0};
\pic (L4) at (L3-end) {link bar generic=100:90pt/0};
\node at (1,5.7) {$i$th calibration point};
\node at (1,5.2) {$(x_i,y_i)$};
\node at (5.7,1.7) {encoder location};
\node at (5.6,1.2) {$(x_0,y_0)$};
\draw[thick,-stealth] (-0.25,0) -- (2,0) node[right]{$x$};
\draw[thick,-stealth] (0,-0.25) -- (0,2) node[left]{$y$};
\draw[dashed] (0,0) -- ++(45:40pt);
\draw[dashed] (0,0) -- ++(-25:40pt);
\draw[dashed] (L1-end) -- ++(-25:30pt);
\draw[dashed] (L1-end) -- ++(35:30pt);
\draw[dashed] (L3-end) -- ++(45:30pt);
\draw[dashed] (L3-end) -- ++(100:30pt);
\draw[-stealth] ($(0,0)+(1,0)$) arc (0:45:1);
\draw[-stealth] ($(0,0)+(1.25,0)$) arc (0:-25:1.24);
\draw[-stealth] ($(L1-end)+(0.6797,-0.3170)$) arc (-25:35:0.75);
\draw[-stealth] ($(L3-end)+(0.5303,0.5303)$) arc (45:100:0.75);
\node at (1.3,0.4) {$\theta_{1_i}$};
\node at (1.9,2.5) {$\theta_{2_i}$};
\node at (3.1,-0.9) {$\theta_{2_0}$};
\node at (1.6,-0.27) {$\theta_{1_0}$};
\draw[thick,-stealth] (L2-end) -- (L4-end);
\node at (3.4,2.5) {$d_i$};
\end{tikzpicture}
}
    \caption{Planar 2 degrees-of-freedom arm}
    \label{FIG:planar2DOFarm}
\end{figure}
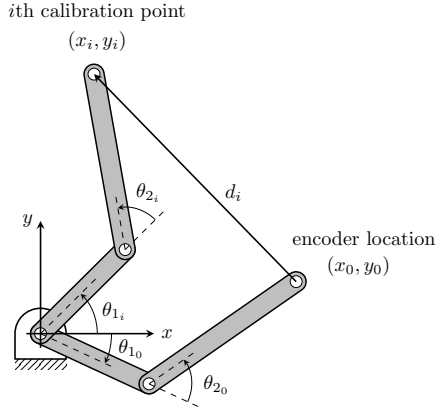
The proposed approach to find different sets of calibration points that only depend on one unknown parameter is here demonstrated in the case of a 2 degrees-of-freedom planar arm, depicted in Figure \ref{FIG:planar2DOFarm}. In this case, the joint coordinates are $\bm{\theta}=[\theta_1,\theta_2]$ while the geometric parameters are the link lengths $\bm{p}=[l_1,l_2]$. Without loss of generality, we assume that we only wish to estimate the joint offset: therefore we define $\bm{e}=[\theta_1, \theta_2]$ and $\bm{\delta e}=[\delta \theta_1, \delta \theta_2]$. During the calibration procedure, the planar arm is first moved to the configuration specified by the joint coordinates $\bm{\theta_0}=[\theta_{1_0},\theta_{2_0}]$ and the encoder is mounted so that the two tips touch when the robot is in this configuration (i.e. the measured distance is null). The planar arm is then moved through a set of $N$ calibration points. Let us consider the $i$th calibration point, specified by the joint coordinates $\bm{\theta_i}=[\theta_{1_i},\theta_{2_i}]$ as depicted in Figure \ref{FIG:planar2DOFarm}. The predicted distance can be easily calculated as
\begin{equation}
    d_i = \sqrt{\Delta x_{i-0}^2+\Delta y_{i-0}^2}= \sqrt{(x_i -x_0)^2+(y_i-y_0)^2}.
    \label{EQ:app_distance}
\end{equation}
This quantity is compared with the distance $\tilde{d}_i$ measured by the draw-wire encoder and the discrepancy between the two is used as a cost function. The coordinates of the $i$th calibration point can be computed through the direct kinematics
\begin{equation}
\begin{cases}
    x_i = f_x(\bm{\theta_i},\bm{p_n}) \\
    y_i = f_y(\bm{\theta_i},\bm{p_n})
    \label{EQ:app_coordinates_i}
\end{cases},
\end{equation}
where $\bm{p_n}$ are the nominal geometric parameters (i.e. nominal link lengths). Since the fixed location of the encoder is specified by the joint coordinates $\bm{\theta_0}$, the coordinates of the encoder location can also be computed through the direct kinematics:
\begin{equation}
\begin{cases}
    x_0 = f_x(\bm{\theta_0},\bm{p_n}) \\
    y_0 = f_y(\bm{\theta_0},\bm{p_n})
    \label{EQ:app_coordinates_0}
\end{cases}.
\end{equation}
However, the joint coordinates are affected by errors: these errors influence both the coordinates of the $i$th calibration point and the encoder coordinates. To evaluate the contribution of the errors on $\theta_1$ and $\theta_2$ on the $i$th calibration point and encoder coordinates, Equations \eqref{EQ:app_coordinates_i} and \eqref{EQ:app_coordinates_0} can be expressed as a Taylor expansions truncated at the first order:
\begin{equation}
\begin{cases}
    x_i = f_x(\bm{\theta_i},\bm{p_n})+\frac{\partial f_x(\bm{\theta},\bm{p})}{\partial \theta_1} \Bigr|_{\substack{\bm{\theta}=\bm{\theta_i}\\ \bm{p}=\bm{p_n}}}
    \delta \theta_1 
    +\frac{\partial f_x(\bm{\theta},\bm{p})}{\partial \theta_2} \Bigr|_{\substack{\bm{\theta}=\bm{\theta_i}\\ \bm{p}=\bm{p_n}}}
    \delta \theta_2 \\
    y_i = f_y(\bm{\theta_i},\bm{p_n})+\frac{\partial f_y(\bm{\theta},\bm{p})}{\partial \theta_1} \Bigr|_{\substack{\bm{\theta}=\bm{\theta_i}\\ \bm{p}=\bm{p_n}}}
    \delta \theta_1 
    +\frac{\partial f_y(\bm{\theta},\bm{p})}{\partial \theta_2} \Bigr|_{\substack{\bm{\theta}=\bm{\theta_i}\\ \bm{p}=\bm{p_n}}}
    \delta \theta_2
\end{cases}
\label{EQ:app_coordinates_1_der}
\end{equation}
\begin{equation}
\begin{cases}
    x_0 = f_x(\bm{\theta_0},\bm{p_n})+\frac{\partial f_x(\bm{\theta},\bm{p})}{\partial \theta_1} \Bigr|_{\substack{\bm{\theta}=\bm{\theta_0}\\ \bm{p}=\bm{p_n}}}
    \delta \theta_1 
    +\frac{\partial f_x(\bm{\theta},\bm{p})}{\partial \theta_2} \Bigr|_{\substack{\bm{\theta}=\bm{\theta_0}\\ \bm{p}=\bm{p_n}}}
    \delta \theta_2 \\
    y_0 = f_y(\bm{\theta_0},\bm{p_n})+\frac{\partial f_y(\bm{\theta},\bm{p})}{\partial \theta_1} \Bigr|_{\substack{\bm{\theta}=\bm{\theta_0}\\ \bm{p}=\bm{p_n}}}
    \delta \theta_1 
    +\frac{\partial f_y(\bm{\theta},\bm{p})}{\partial \theta_2} \Bigr|_{\substack{\bm{\theta}=\bm{\theta_0}\\ \bm{p}=\bm{p_n}}}
    \delta \theta_2
\end{cases}.
\label{EQ:app_coordinates_0_der}
\end{equation}
To simplify the notation, let us call $(x_{i,th},y_{i,th})=(f_x(\bm{\theta_i},\bm{p_n}),f_y(\bm{\theta_i},\bm{p_n}))$ and $(x_{0,th},y_{0,th})=(f_x(\bm{\theta_0},\bm{p_n}),f_y(\bm{\theta_0},\bm{p_n}))$ the theoretical coordinates of the two points (i.e. evaluated with nominal values of the parameters). Let us also define $\Delta x_{i-0,th}=x_{i,th}-x_{0,th}$, $\Delta y_{i-0,th}=y_{i,th}-y_{0,th}$  and the three following matrices:
\begin{equation}
    \bm{\Phi_i} = \begin{bmatrix}
        \partial f_x/\partial \bm{e} \\
        \partial f_y/\partial \bm{e}
    \end{bmatrix} _{\substack{\bm{\theta}=\bm{\theta_i}\\ \bm{p}=\bm{p_{n}}}}=\begin{bmatrix}
        \frac{\partial f_x(\bm{\theta},\bm{p})}{\partial \theta_1} & \frac{\partial f_x(\bm{\theta},\bm{p})}{\partial \theta_2} \\
        \frac{\partial f_y(\bm{\theta},\bm{p})}{\partial \theta_1} & \frac{\partial f_y(\bm{\theta},\bm{p})}{\partial \theta_2}
    \end{bmatrix}_{\substack{\bm{\theta}=\bm{\theta_i}\\ \bm{p}=\bm{p_{n}}}}=
    \begin{bmatrix}
        \Phi_{i_{11}} & \Phi_{i_{12}} \\
        \Phi_{i_{21}} & \Phi_{i_{22}}
    \end{bmatrix} 
\end{equation}
\begin{equation}
    \bm{\Phi_0} = \begin{bmatrix}
        \partial f_x/\partial \bm{e} \\
        \partial f_y/\partial \bm{e}
    \end{bmatrix} _{\substack{\bm{\theta}=\bm{\theta_0}\\ \bm{p}=\bm{p_{n}}}}=\begin{bmatrix}
        \frac{\partial f_x(\bm{\theta},\bm{p})}{\partial \theta_1} & \frac{\partial f_x(\bm{\theta},\bm{p})}{\partial \theta_2} \\
        \frac{\partial f_y(\bm{\theta},\bm{p})}{\partial \theta_1} & \frac{\partial f_y(\bm{\theta},\bm{p})}{\partial \theta_2}
    \end{bmatrix}_{\substack{\bm{\theta}=\bm{\theta_0}\\ \bm{p}=\bm{p_{n}}}}=
    \begin{bmatrix}
        \Phi_{0_{11}} & \Phi_{0_{12}} \\
        \Phi_{0_{21}} & \Phi_{0_{22}}
    \end{bmatrix} 
\end{equation}
\begin{equation}
    \bm{\Phi_{i-0}}=\bm{\Phi_i}-\bm{\Phi_0}.
\end{equation}
With these definitions, Equations \eqref{EQ:app_coordinates_1_der} and \eqref{EQ:app_coordinates_0_der} can be rewritten as follows:
\begin{equation}
\begin{cases}
    x_i=x_{i,th}+\Phi_{i_{11}}\delta \theta_1+\Phi_{i_{12}} \delta \theta_2 \\
    y_i=y_{i,th}+\Phi_{i_{21}}\delta \theta_1+\Phi_{i_{22}} \delta \theta_2
\end{cases}
\end{equation}
\begin{equation}
\begin{cases}
    x_0=x_{0,th}+\Phi_{0_{11}}\delta \theta_1+\Phi_{0_{12}} \delta \theta_2 \\
    y_0=y_{0,th}+\Phi_{0_{21}}\delta \theta_1+\Phi_{0_{22}} \delta \theta_2
\end{cases}.
\end{equation}
The two quantities $\Delta x_{i-0}$ and $\Delta y_{i-0}$ in Equation \eqref{EQ:app_distance} can be expressed as:
\begin{equation}
\begin{cases}
    \Delta x_{i-0} = \Delta x_{i-0,th}+(\Phi_{i_{11}}-\Phi_{0_{11}})\delta \theta_1+(\Phi_{i_{12}}-\Phi_{0_{12}}) \delta \theta_2 \\
    \Delta y_{i-0} = \Delta y_{i-0,th}+(\Phi_{i_{21}}-\Phi_{0_{21}})\delta \theta_1+(\Phi_{i_{22}}+\Phi_{0_{22}}) \delta \theta_2
\end{cases}.
\end{equation}
The partial Jacobians that appear subtracted from each other can be rewritten as a single matrix. The previous expression can be rewritten as:
\begin{equation}
\begin{cases}
    \Delta x_{i-0} = \Delta x_{i-0,th}+\Phi_{i-0_{11}}\delta \theta_1+\Phi_{i-0_{12}}\delta \theta_2 \\
    \Delta y_{i-0} = \Delta y_{i-0,th}+\Phi_{i-0_{21}}\delta \theta_1+\Phi_{i-0_{22}} \delta \theta_2
\end{cases}.
\end{equation}
Those two quantities appear squared when computing the predicted distance $d_i$ between the robot's end-effector and the encoder. Neglecting the infinitesimal of higher order, we obtain
\begin{equation}
\begin{cases}
    \Delta x_{i-0}^2 = \Delta x_{i-0,th}^2+2\Delta x_{i-0,th} (\Phi_{i-0_{11}}\delta \theta_1+\Phi_{i-0_{12}} \delta \theta_2) \\
    \Delta y_{i-0}^2 = \Delta y_{i-0,th}^2+2\Delta y_{i-0,th} (\Phi_{i-0_{21}}\delta \theta_1+\Phi_{i-0_{22}} \delta \theta_2) 
\end{cases}.
\end{equation}
In Equation \eqref{EQ:app_distance}, $\Delta x_{i-0}^2$ and $\Delta y_{i-0}^2$ are then summed to obtain $d_i$, which can be rewritten as
\begin{equation}
    d_i=\sqrt{\Delta x_{i-0,th}^2+\Delta y_{i-0,th}^2+\delta d_{i,\delta \theta_1}+\delta d_{i,\delta \theta_2}},
\end{equation}
where $\delta d_{i,\delta \theta_1}$ and $\delta d_{i,\delta \theta_2}$ are the errors in the predicted distance due to the discrepancy between the nominal and actual values of $\theta_1$ and $\theta_2$. These two quantities can be computed as
\begin{equation}
    \delta d_{i,\delta \theta_1}=2 (\Delta x_{i-0,th} \Phi_{i-0_{11}}+\Delta y_{i-0,th} \Phi_{i-0_{21}})\delta\theta_1 
    \label{EQ:app_delta_theta1}
\end{equation}
\begin{equation}
    \delta d_{i,\delta \theta_2}=2 (\Delta x_{i-0,th} \Phi_{i-0_{11}}+\Delta y_{i-0,th} \Phi_{i-0_{21}})\delta \theta_2.
    \label{EQ:app_delta_theta2}
\end{equation}
By observing Equation \eqref{EQ:app_delta_theta1} and \eqref{EQ:app_delta_theta2}, it becomes clear that the contribution of the error parameters on the predicted distance may also become null. For instance, if the term $(\Delta x_{i-0,th} \Phi_{i-0_{11}}+\Delta y_{i-0,th} \Phi_{i-0_{21}})$ in \eqref{EQ:app_delta_theta1} is null, an error on $\theta_1$ does not influence the distance $d_i$: in this case, at the $i$th calibration point, the quantity $d_i-\tilde{d}_i$ depends only on the error on $\theta_2$. Equations \eqref{EQ:app_delta_theta1} and \eqref{EQ:app_delta_theta2} can be rewritten in matrix form:
\begin{equation}
    \delta d_{i,\delta \theta_1}=2
    \begin{bmatrix}
        \Delta x_{i-0,th} & \Delta y_{i-0,th}
    \end{bmatrix}
    \bm{\Phi_{{i-0}_1}}\delta \theta_1
    \label{EQ:app_delta_theta1_2}
\end{equation}
\begin{equation}
    \delta d_{i,\delta \theta_2}=2
    \begin{bmatrix}
        \Delta x_{i-0,th} & \Delta y_{i-0,th}
    \end{bmatrix}
    \bm{\Phi_{{i-0}_2}}\delta \theta_2,
    \label{EQ:app_delta_theta2_2}
\end{equation}
where $\bm{\Phi_{{i-0}_1}}$ and $\bm{\Phi_{{i-0}_2}}$ are the first and second column of matrix $\bm{\Phi_{i-0}}$. In both cases, the first row is multiplied by $\Delta x_{i-0,th}$ and the second row is multiplied by $\Delta y_{i-0,th}$. Let us define the vector
\begin{equation}
    \bm{\nu_i}=\begin{bmatrix}
        \Delta x_{i-0,th} & \Delta y_{i-0,th}
    \end{bmatrix},
\end{equation}
and the corresponding unit vector $\bm{\hat{\nu}}_i$, representing the theoretical direction of the wire
\begin{equation}
    \bm{\hat{\nu}}_i=\frac{\bm{\nu_i}}{\| \bm{\nu_i} \|}.
\end{equation}
Equations \eqref{EQ:app_delta_theta1_2} and \eqref{EQ:app_delta_theta2_2} can be rewritten as
\begin{equation}
    \delta d_{i,\delta \theta_1}=2 \| \bm{\nu_i} \| \langle \bm{\Phi_{{i-0}_1}},\bm{\hat{\nu}_i} \rangle \delta \theta_1 = 2 \| \bm{\nu_i} \| \Psi_{i_1} \delta \theta_1
\end{equation}
\begin{equation}
    \delta d_{i,\delta \theta_2}=2 \| \bm{\nu_i} \| \langle \bm{\Phi_{{i-0}_2}},\bm{\hat{\nu}_i} \rangle \delta \theta_2 = 2 \| \bm{\nu_i} \| \Psi_{i_2} \delta \theta_1 .
\end{equation}
The two scalars $\Psi_{i_1}$ and $\Psi_{i_2}$, when compared, quantify the reciprocal influence of $\delta \theta_1$ and $\delta \theta_2$ on the distance $d_i$. They can also be grouped into a single matrix:
\begin{equation}
    \bm{\Psi_i}= \begin{bmatrix}
        \Psi_{i_1} & \Psi_{i_2}
    \end{bmatrix},
\end{equation}
which is the same matrix as in Equation \eqref{EQ:Psi}, obtained in the case of a 2 degrees-of-freedom arm where the only unknown error parameters are the joint offsets. This demonstrates the validity of our approach, which can be extended to manipulators with higher degrees of freedom and with more unknown parameters.




\end{appendices}


\bibliography{sn-bibliography}

\end{document}